%% file: main.tex
\documentclass[12pt]{article}
\usepackage{graphicx}
\usepackage{natbib} %comment out if you do not have the package
\usepackage{url} % not crucial - just used below for the URL 
\usepackage{amsmath}
\usepackage{amssymb}
\usepackage{subcaption}
\usepackage{booktabs}
\usepackage{multirow}
\usepackage{comment}
\usepackage{hyperref}
\usepackage{xcolor}
\usepackage[ruled]{algorithm2e}
\usepackage{algorithmic}
\newcommand{\blind}{0}

\newcommand{\bm}[1]{\mathbf{#1}}

\begin{document}

\def\spacingset#1{\renewcommand{\baselinestretch}%
{#1}\small\normalsize} \spacingset{1} % was 1 

%%%%%%%%%%%%%%%%%%%%%%%%%%%%%%%%%%%%%%%%%%%%%%%%%%%%%%%%%%%%%%%%%%%%%%%%%%%%%%

\if0\blind
{
  \title{\bf Simultaneous Monitoring of Shape and Surface Color via 4D Point Clouds: A Registration-free Approach}
\author{}
\date{}
  \author{Mariafrancesca Patalano\textsuperscript{1}\footnote{Corresponding author. Email: \texttt{mariafrancesca.patalano@phd.unipd.it}\\
  This manuscript is currently under review.}, Giovanna Capizzi\textsuperscript{1}, Kamran Paynabar\textsuperscript{2}
    \hspace{.2cm}\\[1em]
    \textsuperscript{1}{Department of Statistical Sciences, University of Padua} \\
\textsuperscript{2}{School of Industrial and Systems Engineering, Georgia Institute of Technology}\\
    }
    
  \maketitle
} \fi

\if1\blind
{
  \bigskip
  \bigskip
  \bigskip
  \begin{center}
    {\LARGE\bf Title}
\end{center}
  \medskip
} \fi

\bigskip
\begin{abstract}
Advanced manufacturing technologies allow for the production of intricate parts featuring high shape complexity and spatially-varying material composition. Data fusion of point clouds with chromatic attributes provides 4D point clouds, a compact and informative representation that encodes both shape and material information.
In this paper, we present a registration-free framework for \underline{S}imultaneous \underline{M}onitoring of sh\underline{A}pe and \underline{C}olor (SMAC) via 4D point clouds. The proposed framework leverages Laplace-Beltrami operator spectral properties to capture and monitor geometric features and the relationship between shape and surface color. A combined monitoring scheme is proposed to effectively detect shape deformations and color anomalies, along with a spatially-aware post-signal diagnostic procedure to determine the source of change and localize color anomalies. Importantly, neither component relies on registration or mesh reconstruction, eliminating error-prone and computationally expensive preprocessing steps. A Monte Carlo simulation study and a case study on functionally graded materials demonstrate that SMAC achieves effective detection performance, particularly for subtle defects, while providing diagnostic capabilities to identify the source and location of anomalies.

\end{abstract}

\noindent%
\vfill

\newpage
\spacingset{2} % DON'T change the spacing!
\input{introLB}

\input{method}

\input{sim}

\input{conclusion}

\appendix

\section*{Supplementary Material}
The code to reproduce Table 1 is available in an \href{https://anonymous.4open.science/r/SMAC-D2E7/README.md}{anonymous GitHub repository}. The online supplementary material includes additional experiments and time complexity analysis.

\section*{Disclosure of Interest}
The authors declare no conflicts of interest.

\section*{Data Availability Statement}

%The code to reproduce the results will be made publicly available via GitHub. 
The Stanford Bunny point cloud is available at the \href{http://graphics.stanford.edu/data/3Dscanrep/}{Stanford 3D Scanning Repository}.
The point cloud of the bicycle saddle is available at \url{https://www.printables.com/model/449742-bicycle-saddle}.

%\clearpage 

%\input{SATMO/supplementary}

\bibliographystyle{apalike}
\bibliography{ref}

\end{document}

%% file: introLB.tex
\section{Introduction}\label{sec:intro}
\begin{figure}[t]
    \centering
\includegraphics[width=0.9\linewidth]{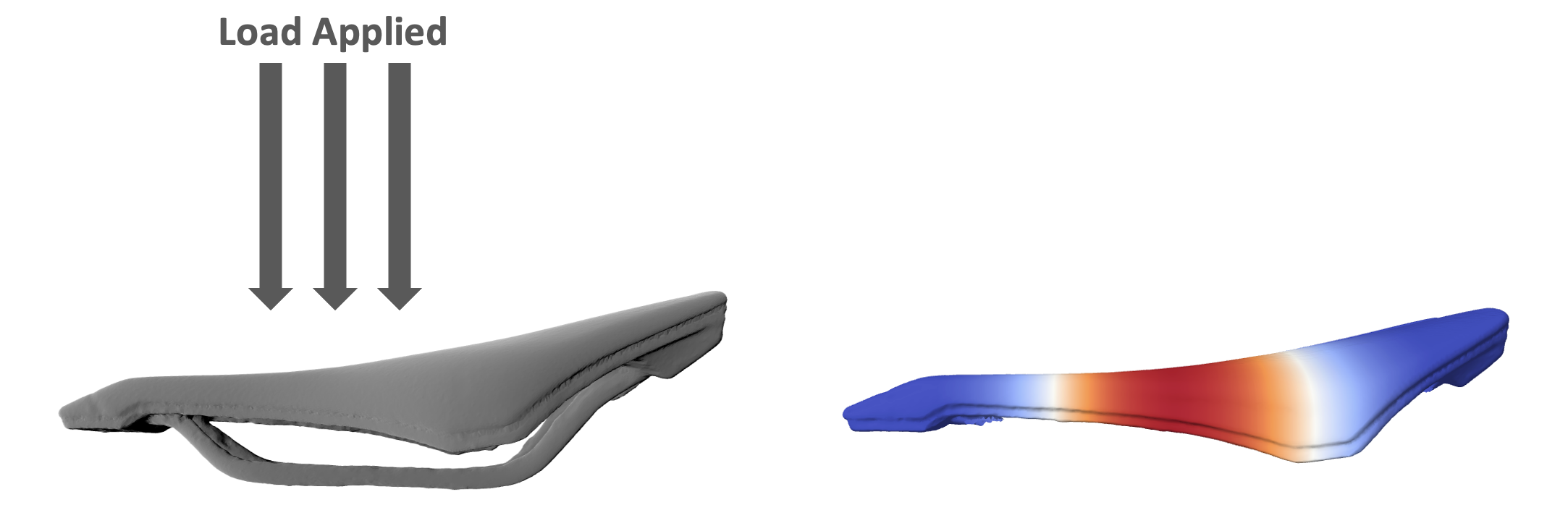}
    \caption{Illustration of the bike seat case study: nominal geometry (left), color-coded representation highlighting the spatially graded rigid (red) and soft (blue) materials (right). 
    }
    \label{fig:bike_sim}
\end{figure}

\begin{figure}[t]
    \centering
    \includegraphics[width=0.6\linewidth]{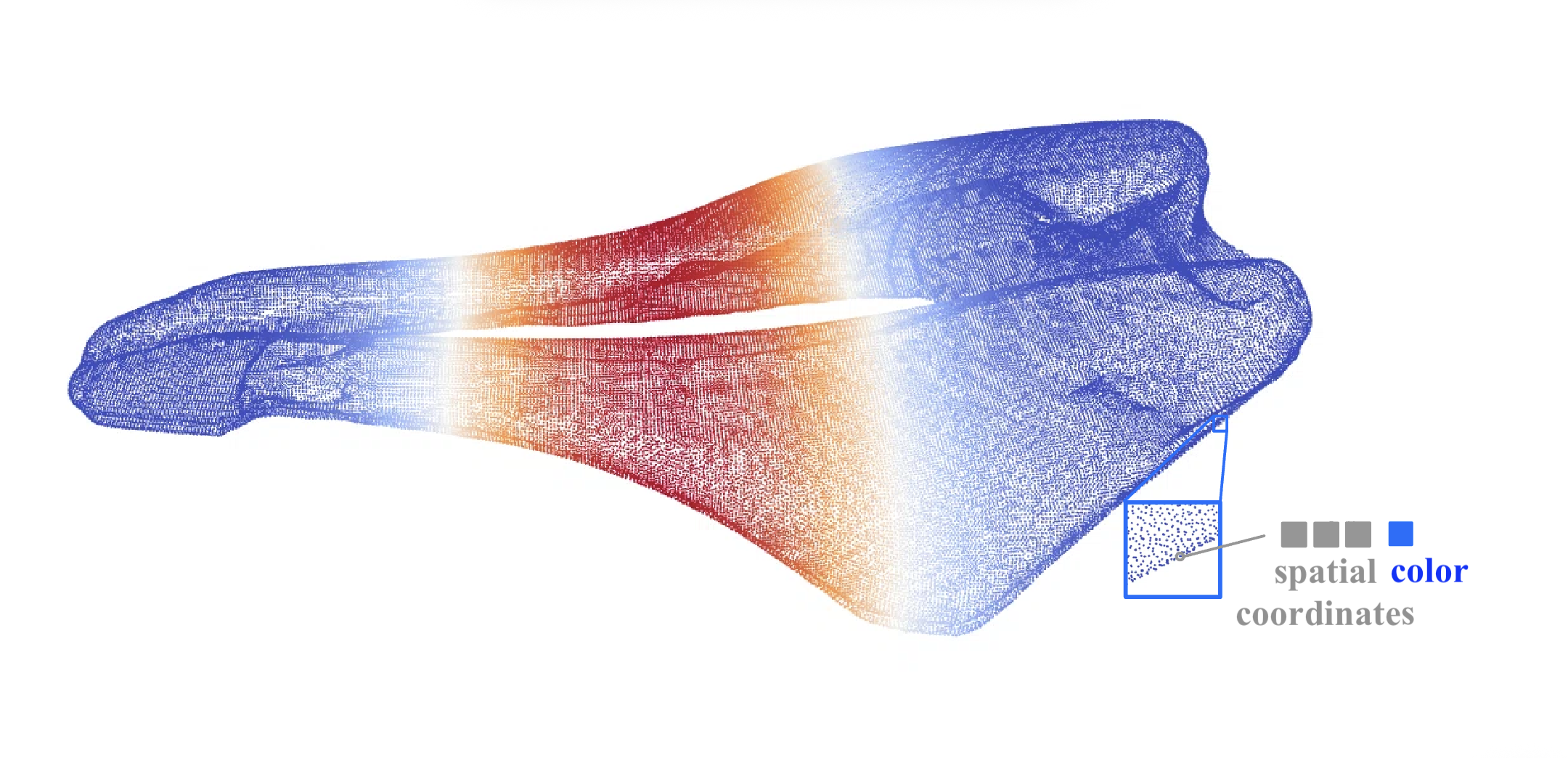}
    \caption{4D point cloud: 3D spatial coordinates are fused with their corresponding chromatic attribute.}
    \label{fig:4dpcd}
\end{figure}

Modern sensing technologies have enabled non-destructive inspection of manufactured parts with intricate geometries, lightweight structures, and topologically optimized designs. 
A prominent application area is the fabrication of Functionally Graded Materials (FGMs), which feature a gradient composition transition between dissimilar materials, enabling the design of new materials with location-dependent mechanical and physical properties \citep{mlfgm2024}. A pragmatic example is the design of a multi-material bicycle seat (Figure \ref{fig:bike_sim}), where rigid and compliant (soft) materials are spatially graded: rigid material is concentrated in regions experiencing excessive deformation under load, while adequately supported contact areas remain predominantly soft to enhance comfort \citep{openvcad}. In multi-material printing, surface color often serves as an indicator of material composition, making chromatic information valuable for quality assessment \citep{rafie2020}.  
In practical applications, it is essential to comprehensively inspect printed components and detect any geometric deviation from the nominal design, as well as any material composition anomalies. The shape inspection is commonly performed ex-situ, namely after the production process, using non-contact metrology systems. %, such as laser scanners and X-ray computed tomography. 
These sensing technologies collect unstructured point cloud data (PCD) and, in contrast to contact-based sensors, enable the inspection of complex geometries. Unstructured PCD exhibit irregular sampling and non-uniform point density, in contrast to structured PCD, in which points are organized in a grid-like structure. However, PCD primarily encode shape information, whereas visible-light and infrared imaging systems capture surface color. These complementary data modalities are often integrated through data fusion, particularly in robotics and autonomous driving applications, to enhance perception and improve environmental understanding (see, e.g., \citealp{li2019, koide2023lidar}), and in manufacturing applications to enable high-precision 3D color reconstruction for stringent inspection tasks \citep{Zhou2025}. By fusing PCD and image data, each 3D point is augmented with its chromatic attribute, resulting in what we refer to as 4D point clouds (Figure \ref{fig:4dpcd}).
The data fusion of point clouds and image data into a 4D point cloud allows for accurate inspection of manufactured components, addressing the low precision, inefficiency, and poor texture measurement of traditional contact-based methods.
Despite recent advances in data fusion, existing statistical process monitoring (SPM) techniques cannot jointly monitor PCD and their point-wise attributes. When monitoring 4D unstructured point clouds several challenges arise: ($i$) high dimensionality: PCD may contain thousands to millions of points, resulting in high-dimensional (HD) data that render conventional monitoring approaches infeasible; ($ii$) varying size: the number of collected points naturally varies between scans, even for identical parts; ($iii$) lack of point-to-point correspondence: unstructured PCD have no natural order, hence point-to-point comparison cannot be performed without additional preprocessing; ($iv$) curved geometry: distances and features must be computed along the manifold surface rather than in the Euclidean space; ($v$) spatial heterogeneity: surface color is a spatially varying attribute and efficient monitoring must account for its inherent location dependence, as global statistics such as mean color are inadequate for detecting localized anomalies.
To address some of these challenges, two preprocessing steps are commonly adopted, namely registration and mesh reconstruction. Registration aligns collected point clouds with a reference geometry (e.g., the computer-aided design (CAD) model), typically via the Iterative Closest Point (ICP) \citep{besl1992} algorithm. 
%The standard algorithm for achieving point cloud registration is Iterative Closest Point (ICP) \citep{besl1992}. 
Nevertheless, ICP has several drawbacks, including a small convergence basin and sensitivity to outliers and partial overlaps. Mesh reconstruction involves constructing a mesh (i.e., a set of vertices, edges, and faces) from the collected PCD to approximate the continuous surface of the manufactured part. However, it is error-prone, potentially introducing holes, artifacts, spurious connections, or even encoding incorrect geometry, thereby affecting monitoring outcomes.
%Mesh reconstruction is another commonly adopted preprocessing step, where a mesh (i.e., a set of vertices, edges, and faces) is constructed to approximate the continuous surface of the manufactured part from the collected PCD. However, mesh reconstruction is error-prone, potentially introducing holes, artifacts, spurious connections, or even encoding incorrect geometry, thereby affecting monitoring outcomes.
Therefore, developing a monitoring framework that detects shape and color anomalies without registration or mesh reconstruction remains an open challenge.

In this paper, we propose a novel registration-free framework for \underline{S}imultaneous \underline{M}onitoring of sh\underline{A}pe and \underline{C}olor (SMAC) via 4D point clouds. To our knowledge, this is the first approach capable of monitoring parts with spatially-varying material composition via 4D point clouds, where chromatic attributes are fused with 3D spatial coordinates.
Unlike existing methods that monitor shape and color separately, SMAC jointly monitors both sources of information within a unified framework. It efficiently handles raw, unaligned, and high-resolution point clouds without requiring mesh reconstruction, registration, or downsampling. 
SMAC is particularly well suited for applications involving complex shapes with spatially heterogeneous material properties, while avoiding false alarms under non-rigid deformations. 
In particular, our approach leverages the Laplace-Beltrami spectral representation to extract hundreds of hierarchically ordered geometric features, capturing global shape descriptors and fine-scale details. The main contributions include: ($i$) the incorporation of point-wise attributes through a function-centric representation that avoids the complexities of explicit surface parameterization while remaining invariant to rotations, translations, and isometric deformations; ($ii$) the design and implementation of a combined monitoring scheme that handles hundreds of features; and ($iii$) a post-signal diagnostic procedure to identify the source of change and localize color anomalies.
SMAC achieves online monitoring of 4D point clouds collected after each part's production. Prior to monitoring, a sample of in-control (IC) point clouds is required, often available in high-volume production, or generated from the CAD model with estimated noise (see, e.g., \citealp{senin2021}) in low-volume applications. 

The remainder of the paper is organized as follows: Section~\ref{sec:relatedwork} introduces useful background concepts and reviews related work. Section~\ref{sec:method} details the proposed methodology, and its performance is assessed through simulation studies in Section~\ref{sec:sim}. Section~\ref{sec:casestudy} shows the application of the proposed approach to a case study, and Section~\ref{sec:conclusion} concludes the paper.

\section{Background and Related Work}\label{sec:relatedwork}
In this section, we first introduce the Laplace--Beltrami operator (LBO) and subsequently review existing literature relevant to our work.

The LBO is the counterpart of the standard Euclidean Laplacian on curved surfaces or manifolds. Given a function $f: \mathcal{M} \rightarrow \mathbb{R}$ defined on a manifold $\mathcal{M}$, LBO is defined as 
\begin{equation}
    \Delta_\mathcal{M} \, f = - \nabla_{\mathcal{M}} \, \cdot \nabla_{\mathcal{M}} \, f
\end{equation}
where $\nabla_{\mathcal{M}}$ and $\nabla_{\mathcal{M}} \, \cdot$ denote the gradient and the divergence operators defined on $\mathcal{M}$. Intuitively, $\Delta_\mathcal{M} \, f$ measures the deviation of the function $f$ from its local average, and characterizes numerous phenomena such as heat diffusion and wave propagation. Although LBO is mathematically defined on a continuous manifold, in practice only a discrete approximation of the underlying manifold is available, typically in the form of a point cloud, mesh, or voxel grid. Therefore, several LBO discretizations have been proposed (see, e.g., \citealp{pinkall1993, belkin2009pcl, Sharp2020tuftedlap}). 
The spectral decomposition of LBO yields an intrinsic coordinate system, given by eigenvalues and eigenfunctions, that depends only on the manifold's intrinsic geometry and is hence invariant to rotations, translations and isometric deformations.  %rather than its embedding in the ambient space (i.e., regardless of rotations, translations, and isometric deformations). 
The spectrum acts as a shape fingerprint, providing an invariant representation for shape identification and comparison \citep{reuter2006cad}. The eigenvalues encode shape information hierarchically, with lower eigenvalues capturing global, coarse-scale geometric features, while higher eigenvalues progressively encode finer local details. The eigenvectors approximate continuous eigenfunctions and provide a function basis over the shape, facilitating the attachment of attributes to surfaces \citep{levy2006}. In this paper, we refer to eigenvectors as eigenfunctions. 

The LBO has been widely adopted beyond the monitoring literature, particularly in computer graphics and medical imaging. LBO spectral properties have been adopted for shape classification and segmentation \citep{Rustamov07, REUTER2009, Lai2009nodalcounts}, and salient point detection \citep{Hu2009Salient,niu2019salient}. In medical imaging, LBO eigenfunctions have been used for surface reconstruction \citep{Shi2010Reconstruction}, shape modeling \citep{kim2012sparse}, and smoothing on cortical manifolds \citep{qiu2006}.

In SPM literature, several approaches have been proposed for assessing the geometric accuracy of 3D manufactured parts. These approaches can be broadly categorized into deviation-based and registration-free methods. Deviation-based methods detect geometric deviations by analyzing distances between the manufactured part and the nominal geometry (i.e., the CAD model). One class of deviation-based approaches subdivides the collected point cloud into voxels or regions of interest (ROIs) and performs region-wise comparisons \citep{huang2018, stankus2019, colosimo2022}. However, defining ROIs or regular voxels can be challenging for parts featuring complex geometries. Alternatively, other methods analyze the distribution of point-to-point distances between the collected and nominal point clouds \citep{wells2013, scimone2022}; though, spatial information is often lost when summarizing point-to-point distances. The main drawback of deviation-based approaches is their reliance on accurate registration to establish point-wise or region-wise correspondence. As a result, their performance is highly sensitive to alignment errors and registration can become computationally expensive for large PCD.
Registration-free methods circumvent the need for spatial alignment by leveraging intrinsic geometric properties, with recent approaches adopting LBO  eigenvalues for monitoring meshes \citep{zhao2021, zhao2022FEM} and unstructured point clouds \citep{rfm}.
However, existing methods are designed exclusively for geometric inspection and are limited in their ability to model point-wise attributes such as surface color. 
%In addition, LBO eigenfunctions have not been exploited for monitoring, despite their capability to attach attributes to the surface and capture localized shape differences.

Separately, image-based SPM methods have been extensively 
developed for monitoring surface appearance and texture, including 
model-based approaches for stochastic textured surfaces \citep{bui2018a, bui2018}, techniques based on matrix and tensor decompositions \citep{yan2015, yanSSD, colosimostpca, yang2023}, and convolutional neural networks (see, e.g., \citealp{wang2018, liu2019, cui2020}). 
Although effective in identifying color and texture anomalies, these methods operate exclusively on 2D representations and therefore cannot adequately model the underlying 3D geometry. Consequently, image-based methods are not directly applicable to PCD and fail to detect geometric deformations in 3D printed parts. 
%Our work addresses this gap by enabling the simultaneous monitoring of shape and color in 4D point clouds without requiring registration or mesh reconstruction.
Existing work reveals a clear gap in the literature, as registration-free SPM methods are limited to geometric inspection, 
while image-based methods detect color and texture anomalies. As a result, no existing method directly enables the simultaneous monitoring of shape and color in 4D point clouds.

%% file: method.tex
\section{Methodology}\label{sec:method}

\subsection{Modeling and Feature Learning}
Let $\mathbf{X}_i = \{\mathbf{S}_i, \boldsymbol{y}_i\} \in \mathbb{R}^{n_i \times 4}$ denote the 4D point cloud associated with the $i-$th sample, where $\mathbf{S}_i \in \mathbb{R}^{n_i \times 3}$ contains the 3D spatial coordinates and $\boldsymbol{y}_i \in \mathbb{R}^{n_i}$ represents the surface color values. We adopt the discrete LBO proposed by \cite{Sharp2020tuftedlap} and compute the stiffness matrix $\mathbf{L}_i \in \mathbb{R}^{n_i \times n_i}$ and the mass matrix $\mathbf{M}_i\in \mathbb{R}^{n_i \times n_i}$ starting from the point cloud $\mathbf{S}_i \in \mathbb{R}^{n_i \times 3}$. In particular, $\mathbf{L}_i$ encodes the local geometric structure by weighting relationships between neighboring points, while $\mathbf{M}_i$ is a diagonal matrix whose entries represent the surface area locally attributed to each point.
Then, the following generalized eigenvalue problem is solved for the first $k$ eigenpairs
\begin{equation}\label{eq:gep}
    \mathbf{L}_i \boldsymbol{u}_{i,j} = \lambda_{i,j}\mathbf{M}_i\boldsymbol{u}_{i,j} 
\end{equation}
for $j = 1, \dots, k$.
From Eq.~(\ref{eq:gep}), for the $i-$th point cloud, we obtain the lower spectrum $\boldsymbol{\lambda}_i \in \mathbb{R}^{k}$ and the eigenfunctions $\mathbf{U}_i = \left[\boldsymbol{u}_{i,1}, \dots,\boldsymbol{u}_{i,k} \right] \in \mathbb{R}^{n_i \times k}$. The first eigenvalue is zero by definition and is therefore omitted from subsequent analyses.

To jointly monitor shape and texture, we leverage the eigenfunctions as a functional basis defined over the object and use them to attach attributes to the surface, following the function-centric view of \citet{levy2006}.
Therefore, we learn the relationship between the shape properties and the observed color values, and define
\begin{equation}\label{eq:lm}
    \boldsymbol{y}_i = \mathbf{U}_{i} \, \boldsymbol{\beta}_i + \boldsymbol{\varepsilon}_i
\end{equation}
\begin{figure}[t]
    \centering
    \includegraphics[width=1.0\linewidth]{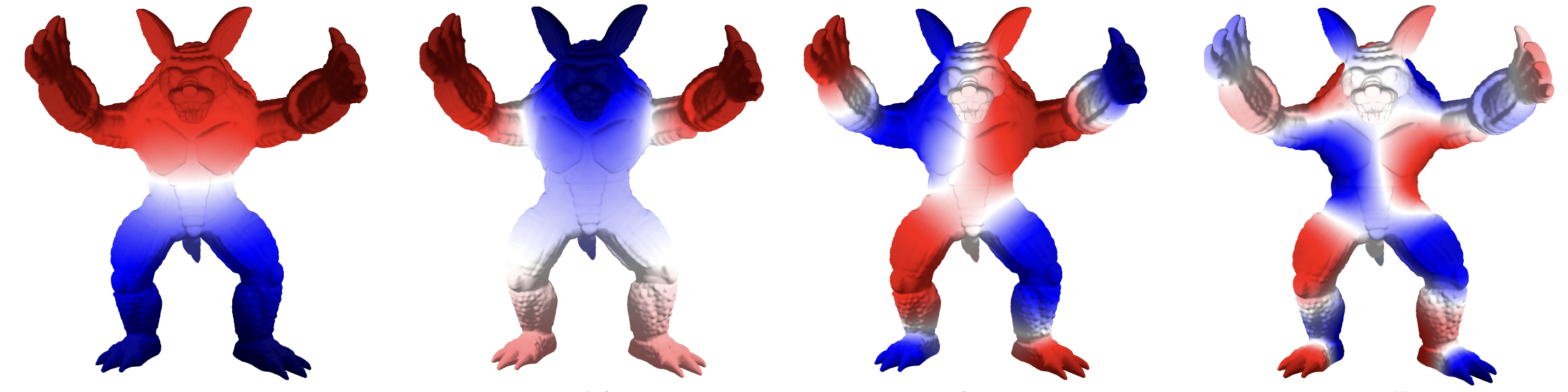}
    \caption{Adaptive shape partitioning by LBO eigenfunctions. From left to right: 2nd, 5th, 10th, and 20th eigenfunctions. Color and its intensity indicate sign and magnitude, where red and blue represent positive and negative values, respectively.}
    \label{fig:partitions}
\end{figure}
where $\boldsymbol{\beta}_i\in \mathbb{R}^k$ is an unknown vector of coefficients and $\boldsymbol{\varepsilon}_i \in \mathbb{R}^{n_i}$ is an error term accounting for residual variability. No specific distributional assumption is imposed on $\boldsymbol{\varepsilon}_i$, the lower spectrum, or the color values. In Eq.~(\ref{eq:lm}), we use eigenfunctions as features, since each eigenfunction adaptively partitions the shape into physically meaningful domains, capturing symmetries. 
These partitions, known as nodal domains, increase in number with eigenfunction order: the number of nodal domains of the $j-$th eigenfunction is at most $j$ \citep{courant1989}.  For instance, Figure \ref{fig:partitions} illustrates that the shape is partitioned into increasingly fine-grained domains, from broad regions subdividing the lower and upper body to localized regions at higher orders capturing fingers and ears.
It is important to note that eigenfunctions are orthogonal with respect to the $\mathbf{M}_i$-inner product, that is $\langle \boldsymbol{u}_j, \boldsymbol{u}_\ell\rangle_{\mathbf{M}_i} = \boldsymbol{u}_j^\top \mathbf{M}_i \boldsymbol{u}_\ell = 0, \, j \neq \ell$, and $j, \ell = 1, \dots, k$ \citep{Rustamov07, REUTER2009}. Therefore, for the $i-$th 4D point cloud, we minimize a mass-weighted error 
\begin{equation}\label{eq:minimize}
    \hat{\boldsymbol{\beta}}_i = \underset{\boldsymbol{\beta}_i}{\text{argmin}} \, \| \boldsymbol{y}_i - \mathbf{U}_{i} \boldsymbol{\beta}_i\|_{\mathbf{M}_i}^2
\end{equation}
and consequently $\hat{\boldsymbol{\beta}}_i$ is given by
\begin{equation}\label{eq:coeff}
   \hat{\boldsymbol{\beta}}_i = (\mathbf{U}_{i}^\top \mathbf{M}_i \mathbf{U}_i)^{-1} \mathbf{U}_{i}^\top \mathbf{M}_i \boldsymbol{y}_i
\end{equation}
\begin{figure}[t]
    \centering
    \includegraphics[width=0.5\linewidth]{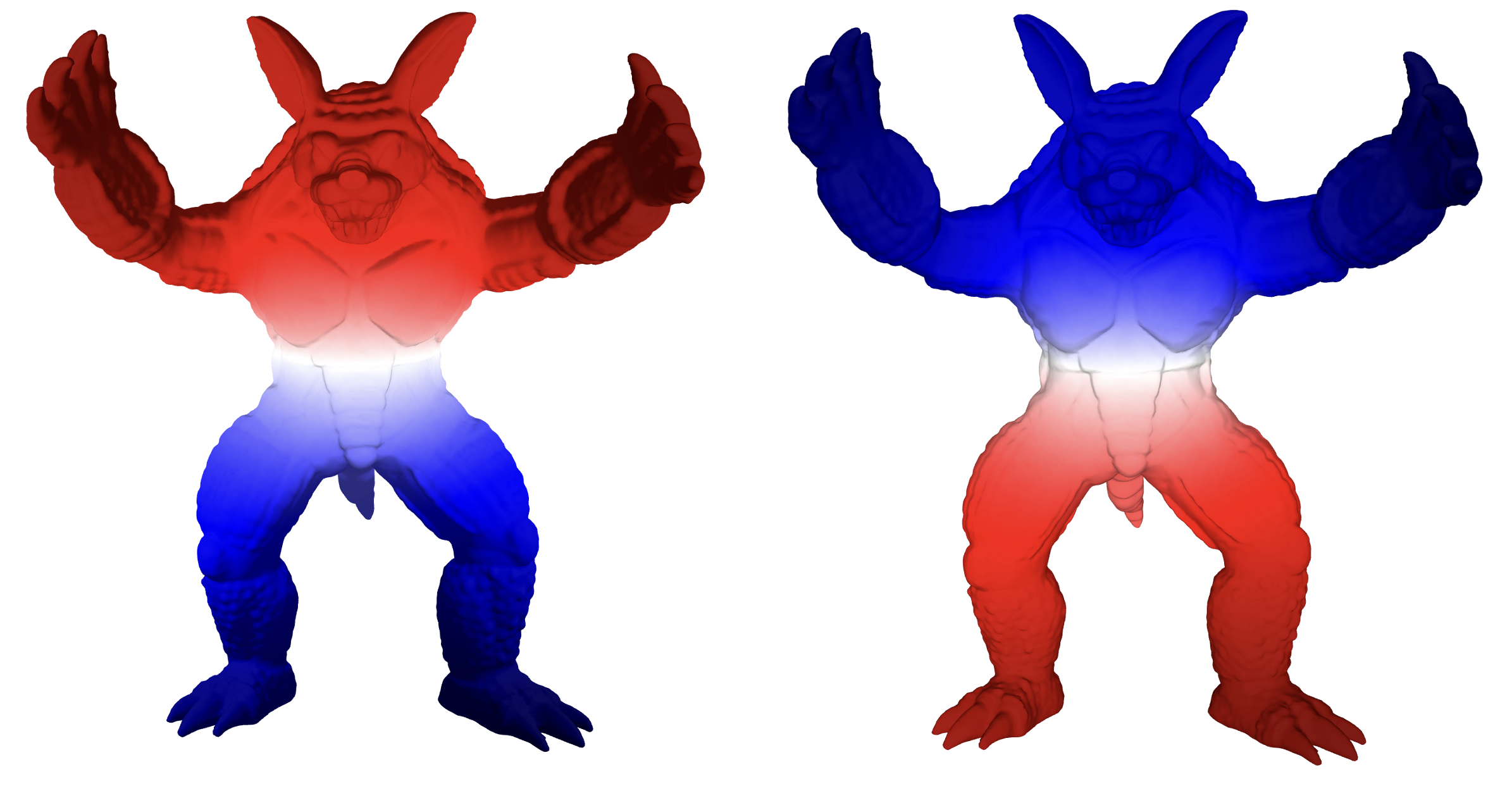}
    \caption{Sign ambiguity in LBO eigenfunctions. Both visualizations represent the same eigenfunction (i.e., the 2nd) with opposite signs.}
    \label{fig:basisamb}
\end{figure}
where $\mathbf{U}_{i}^\top \mathbf{M}_i \mathbf{U}_i = \mathbf{I}_k$ because of orthogonality. Following the standard approach in SPM, we can estimate the vector of coefficients, as in Eq.~(\ref{eq:coeff}), for each point cloud in the reference sample, and average the estimated coefficient vectors. However, eigenfunctions face two challenges: sign ambiguity and basis ambiguity. Sign ambiguity means that an eigenfunction remains valid if its sign is flipped (see Figure \ref{fig:basisamb}). Basis ambiguity arises when an eigenvalue has multiplicity greater than one, allowing any orthogonal basis in the subspace spanned by the corresponding eigenfunctions to serve as a valid set of eigenfunctions \citep{ma2023}. Eigenvalues with multiplicity greater than one may occur in shapes featuring exact symmetries \citep{patane2017}. Nevertheless, in real-world applications, exact symmetries are uncommon, since manufactured parts are never perfectly symmetrical and are subject to even minimal amounts of noise. Hence, in our setting, it is necessary to only address sign ambiguity. To this end, we monitor the absolute value of the coefficients, since for detection purposes, their magnitude is indicative of potential anomalies in the surface color.

\subsection{Monitoring}
In this article, we design a combined monitoring scheme to simultaneously monitor shape and color, which triggers an alarm whenever at least one of the two individual control schemes signals. For representative methods of combined control charts see \cite{tsung2007, han2007, Reynolds2008}.
In the following, the two individual schemes are referred to as the shape and color charts, respectively. 

For each $i-$th point cloud, two feature vectors are available from the previous step: the lower spectrum $\boldsymbol{\lambda}_i \in \mathbb{R}^{k-1}$ and the vector of coefficients in absolute value, denoted $\tilde{\boldsymbol{\beta}}_i = |\hat{\boldsymbol{\beta}}_i| \in \mathbb{R}^k$.
The feature vectors are standardized using their respective sample means and covariance matrices, estimated using the reference sample. In the following, for ease of notation, we denote the standardized observations as $\boldsymbol{\lambda}_i$ and $\tilde{\boldsymbol{\beta}}_i$. In both schemes, we adopt the $L_2$ norm as a control statistic, resulting in $s_i = \|\boldsymbol{\lambda}_i\|_2^2$ and $c_i = \|\tilde{\boldsymbol{\beta}}_i\|_2^2$. 
Both statistics are monitored using univariate cumulative sum (CUSUM) charts \citep{pagecusum}. The CUSUM chart is designed to detect persistent shifts from the target value by accumulating deviations over time. For an observation $x_i$, the one-sided upper CUSUM is defined as
\begin{equation}\label{eq:cusum}
    C_{x,i}^+ = \max \left(0, \, C_{x,i-1}^+ + \frac{x_i - \mu_x}{\sigma_x} - k_x\right),
\end{equation}
where $\mu_x$ and $\sigma_x$ are the IC mean and standard deviation, $C_{x,0}^+ = 0$, and $k_x > 0$ is a reference value. An alarm is triggered when $C_{x,i}^+ > h_x$, where $h_x$ is the control limit.

At the $i-$th time step, $s_i$ and $c_i$ are computed and used to update the CUSUM statistics $C_{s,i}^+$ and $C_{c,i}^+$ of the shape and color charts, respectively. The parameters of the shape and color charts are $(\mu_s, \sigma_s, k_s, h_s)$ and $(\mu_c, \sigma_c, k_c, h_c)$, respectively. 
The combined scheme signals an out-of-control (OC) condition at time $i$ if $C_{s,i}^+ > h_s$ or $C_{c,i}^+ > h_c$.

No assumptions are imposed on the distribution of the feature vectors or on the direction, pattern, or shift magnitude.
Notably, the two schemes are coupled: the color chart is sensitive to changes in both color-shape relationship and shape alone because the estimated coefficients $\hat{\boldsymbol{\beta}}_i$ depend on the eigenfunctions, which in turn depend on the shape. Although this coupling enables the combined scheme to quickly detect anomalies, an OC signal from the color chart may reflect either type of change. To resolve this ambiguity, we develop a post-signal diagnostic procedure in Section~\ref{sec:diagnostics}.

\subsection{Post-signal Diagnostic}\label{sec:diagnostics}
After an alarm is triggered by the color chart, the post-signal diagnostic aims to determine whether the deviation originates from a shape change alone or from a joint shape and color change. To this end, we propose a formal test on the color component, which resolves the aforementioned ambiguity and enables localization of a potential color defect on the 3D object. 

Our objective is to test, at a point-wise level, whether the color significantly deviates from the nominal values. To apply the test, we map all color values to the common reference space of the CAD model using functional maps \citep{Ovsjanikov2012}, which transport functions between shapes without requiring explicit point or region correspondences. 

\subsubsection{Computation of Functional Maps}
Point clouds naturally differ in size and lack a consistent point-to-point correspondence. For this reason, we rely on functional maps to transport the color information to a common reference domain defined by the CAD model.  
Given two sets of point clouds containing IC and detected OC samples, respectively, we compute for each point cloud the functional map $\mathbf{C} \in \mathbb{R}^{k \times k}$ between the CAD model $\mathbf{S}_0 \in \mathbb{R}^{n_0 \times 3}$ and a test shape $\mathbf{S}_1 \in \mathbb{R}^{n_1 \times 3}$ by solving 
\begin{equation}\label{eq:fm}
\begin{split}
&\underset{{\mathbf{C}}}{\text{argmin} } \, \|\mathbf{CA}_0 - \mathbf{A}_1\|^2 + \eta \|\mathbf{\Lambda}^{\mathbf{S}_1} \mathbf{C} -\mathbf{C} \mathbf{\Lambda}^{\mathbf{S}_0}\|^2 =\\
&\underset{{\mathbf{C}}}{\text{argmin} } \, \|\mathbf{CA}_0 - \mathbf{A}_1\|^2 + \eta \sum_{i,j} {c}_{ij}^2 (\lambda_i^{\mathbf{S}_1} - \lambda_j^{\mathbf{S}_0})^2
\end{split}
\end{equation}
where $\mathbf{\Lambda}^{\mathbf{S}_0}$ and $\mathbf{\Lambda}^{\mathbf{S}_1}$ are diagonal matrices of eigenvalues of LBO, and $\eta$ is a regularization parameter that penalizes deviation from Laplacian commutativity (i.e., encourages each eigenfunction of $\bm{S}_0$ to be mapped to the eigenfunction of $\bm{S}_1$ with the closest eigenvalue). % thereby encouraging the map to preserve intrinsic geometry (i.e., promoting near-isometry). 
%The first term in Eq.\,(\ref{eq:fm}) encourages descriptor preservation, with $\mathbf{A}_0 = \mathbf{U}_0^\top \mathbf{M}_0  \mathbf{D}_0 \in \mathbb{R}^{k \times p}$ (and equivalently for $\mathbf{A}_1$). Here, $\mathbf{D}_0 \in \mathbb{R}^{n_0 \times p}$ and $\mathbf{D}_1 \in \mathbb{R}^{n_1 \times p}$ contain $p$ point-wise descriptors, namely heat kernel signature and wave kernel signature \citep{HKS, WKS}.
The first term of Eq.\,(\ref{eq:fm}) includes $\mathbf{A}_0 = \mathbf{U}_0^\top \mathbf{M}_0  \mathbf{D}_0 \in \mathbb{R}^{k \times p}$ and $\mathbf{A}_1 = \mathbf{U}_1^\top \mathbf{M}_1  \mathbf{D}_1 \in \mathbb{R}^{k \times p}$, where $\mathbf{D}_0 \in \mathbb{R}^{n_0 \times p}$ and $\mathbf{D}_1 \in \mathbb{R}^{n_1 \times p}$ contain $p$ point-wise descriptors, namely heat kernel signature and wave kernel signature \citep{HKS, WKS}, which capture multi-scale, isometry-invariant geometric information at each point. Since corresponding points on near-isometric shapes should carry similar descriptor values, this term encourages $\bm{C}$ to map descriptors of $\bm{S}_0$ as closely as possible onto those of $\bm{S}_1$ (i.e., descriptor preservation).

\subsubsection{Mapping}
Denote as $\mathbf{C}_i \in\mathbb{R}^{k \times k}$ the functional map between the CAD model and the $i-$th point cloud. The map $\mathbf{C}_i$ is used to transport the coefficients $\hat{\boldsymbol{\beta}}_i$ to the CAD model eigenspace, obtaining $\boldsymbol{\gamma}_i = \mathbf{C}_i^\top \hat{\boldsymbol{\beta}}_i$, with $\boldsymbol{\gamma}_i \in \mathbb{R}^k$. Then, we use the vector of transported coefficients $\boldsymbol{\gamma}_i$ to reconstruct the color value of the $i-$th point cloud on the reference shape by projection onto its eigenbasis
\begin{equation}\label{eq:texture}
    \boldsymbol{y}_0^{(i)} = \mathbf{U}_0 \boldsymbol{\gamma}_i\,, \, \quad  \boldsymbol{y}_0^{(i)} \in \mathbb{R}^{n_0}
\end{equation}
Repeating this procedure for all the point clouds in the IC and OC sets yields color vectors of the same dimension $n_0$ defined on the reference shape (i.e., the CAD model). Sharing a common reference space enables point-wise comparison even for point clouds of varying size.\\

\subsubsection{Spatially-aware Point-wise Test}
Having mapped all color values to the common reference space, we perform a two-sample test at each point to assess whether the color values of the OC samples significantly deviate from those of the IC samples. To account for spatial dependence between neighboring points, we employ a nonparametric cluster-based permutation procedure originally developed for neuroimaging data \citep{maris2007, smith2009}.
Specifically, an adjacency matrix is computed from $\mathbf{S}_0 \in \mathbb{R}^{n_0 \times 3}$ using k-nearest neighbors to define the spatial vicinity between points. For each point $q = 1, \dots, n_0$, the F-statistic is computed to compare IC and OC groups. To avoid arbitrary threshold selection, we apply the threshold-free cluster enhancement \citep{smith2009}, which leverages the adjacency structure to transform point-wise statistics into cluster-enhanced statistics by integrating over spatially contiguous regions at multiple threshold levels.  Statistical significance is assessed via permutation testing: IC and OC labels are randomly permuted and the maximum cluster-enhanced statistic across all points is computed for each permutation. This procedure controls the family-wise error rate (FWER) at level $\alpha$, ensuring strong control over false positives across all $n_0$ simultaneous tests. Points or clusters of points with permutation-based p-values below $\alpha$ are declared significant, indicating spatial regions where color deviates from nominal values.

\subsection{Design of the Chart and Choice of Tuning Parameters}
The design of the SMAC monitoring scheme involves selecting the number of eigenpairs $k$ and the size of the reference sample, used for estimating the parameters to standardize eigenvalues and coefficients, and the chart-specific parameters $(\mu_s, \sigma_s, k_s, h_s)$ and $(\mu_c, \sigma_c, k_c, h_c)$. The number of eigenpairs $k$ is selected using the elbow-rule procedure proposed by \cite{rfm} and summarized in Section S4 of the Supplementary Material. 
We subdivide the reference sample into three subsets of size $m_1$, $m_2$ and $m_3$. The first subset is used to compute the $k-1$ and $k-$ dimensional means and covariance matrices to standardize the eigenvalues and coefficients, respectively. The second subset is used to estimate $(\mu_s, \sigma_s)$ and $(\mu_c, \sigma_c)$, while the last is used to compute $h_s$ and $h_c$. Based on our empirical findings, we recommend $m_1 = 600$ and $m_2 = 300$. The value of $m_1$ should be chosen relative to the number of eigenpairs $k$, as a reliable estimation of the mean and covariance matrix requires a sufficiently large sample size relative to the dimension. As a practical guideline, we recommend inspecting the difference $\|\hat{\mathbf{\Sigma}} -   \mathbf{I}_k \|_F$, where $\hat{\mathbf{\Sigma}}$ is the estimated covariance matrix of the standardized observations and $\| \cdot \|_F$ is the Frobenius norm. %The difference should be small once $m_1$ is sufficiently large. 
Notably, the choice of $k$, and consequently of $m_1$, may ultimately be guided by the data available for calibration and the monitoring requirements of the specific application, since a smaller $k$ reduces the number of samples required for calibration, but may reduce detection accuracy.
The value $m_2 = 300$ suffices for accurate estimation of the mean and variance of the monitoring statistics $s_i$ and $c_i$. 
A sensitivity analysis for $m_3$ is provided in Section \ref{sec:sim}. To compute $h_s$ and $h_c$, we adapt the bootstrap-assisted bisection algorithm of \cite{Zago2025} to accommodate our combined monitoring scheme. The CUSUM reference values $k_s$ and $k_c$ are set according to standard guidelines, with values commonly in $[0.05, 0.2]$; smaller values detect subtle defects more effectively, while larger values are more sensitive to severe defects.

%% file: sim.tex
\section{Simulation Study}\label{sec:sim}
In this section, we evaluate the performance of SMAC on localized defects, which are more challenging to detect than global or extended anomalies. We consider the Stanford Bunny point cloud as the nominal design, a standard benchmark in computer graphics and geometry processing. The Bunny has approximate dimensions of 0.15 mm in height and width and consists of 8,146 points. A chromatic attribute is attached to each point, varying smoothly along the Z-axis, with values ranging between 0 (yellow) and 10 (purple). Noisy point clouds are generated with varying sizes, isotropic noise $N_3(\bm{0}, \tau_s^2 \, \bm{I})$ with $\tau_s = 0.0001$, and chromatic noise $N(0,\tau_c^2)$ with $\tau_c = 0.01$. In particular, in Sections \ref{sec:sim-icoc} and \ref{sec:sim-ps} noisy point clouds have size ranging between 8,100 and 8,140 points, while in Section \ref{sec:sim-rob} their size ranges between 6,480 and 8,140. Performance is assessed using the average run length (ARL), where the run length is defined as the number of observations (here, point clouds) before an alarm is triggered. Results are reported in terms of average ARL (AARL), along with its standard deviation, across 30 Monte Carlo replications. The IC ARL, denoted as $\text{ARL}_0$, is set to $100$ and the maximum run length is set to 4,000. 
Three competitive methods are considered for comparison: ISOMAP \citep{tenenbaum2000}, Locally Linear Embedding (LLE) \citep{roweis2000lle}, and Graph Laplacian (GL) \citep{chung1997}. ISOMAP and LLE are established manifold learning methods, while GL is included due to its analogy with LBO (see \citealp{chung1997}). %Despite being applicable for monitoring 4D point clouds, the eigenfunctions, computed from their spectral decomposition, do not provide meaningful shape partitions. 
\begin{figure}[t]
    \centering
    \includegraphics[width=0.7\linewidth]{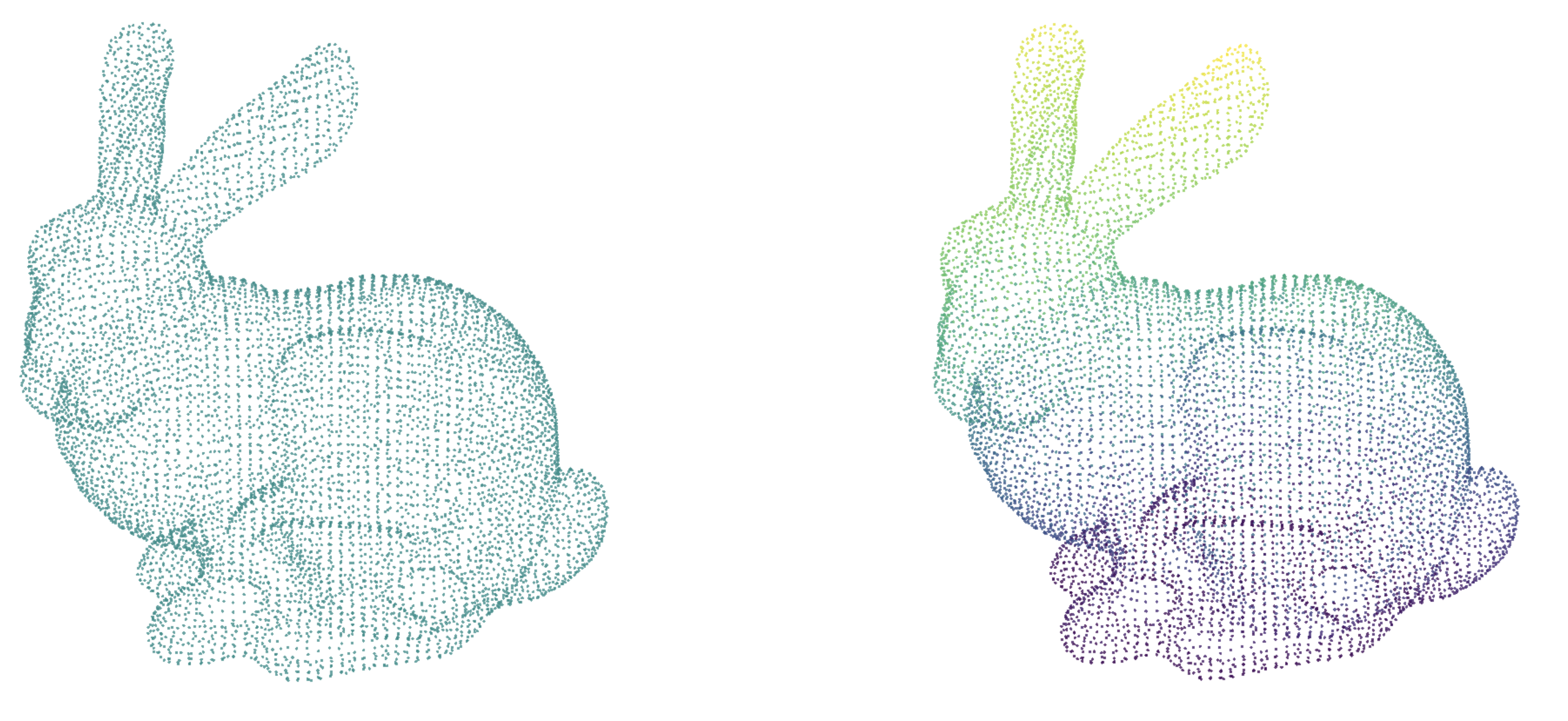}
    \caption{Stanford Bunny point cloud (left) with chromatic attribute (right).}
    \label{fig:bunny_plot}
\end{figure}
These methods are applied directly to 4D point clouds, in contrast to SMAC that processes spatial and chromatic information separately.  
In fact, despite being applicable to 3D and 4D point clouds without requiring registration or mesh reconstruction, the eigenfunctions computed from their spectral decomposition do not provide meaningful shape partitions. Hence, eigenvalues serve as features: eigenvalues of the GL matrix, and, for ISOMAP and LLE, eigenvalues are computed via classical multidimensional scaling applied to pairwise distances in the embedding space. Then, the eigenvalues are standardized and their $L_2$ norm is provided to a single CUSUM chart. 
Throughout all experiments, we use 30 nearest neighbors for constructing LBO, GL, ISOMAP, and LLE, following the recommendations in \cite{Sharp2020tuftedlap}, and 1,000 bootstrap samples in each Monte Carlo replication. The number of eigenvalues is set to 100 for SMAC and GL (i.e., $k = 101$), and to the embedding dimension for ISOMAP and LLE. The CUSUM reference values $k_s$ and $k_c$ are set to 0.05 (additional results for different values of $k_s$ and $k_c$ are reported in the Supplementary Material Section S2). All tuning parameters are fixed across all simulations and selected before comparing the methods, independently of their relative performance, so that their choice does not favor any particular method. For brevity, we report results for a single configuration.

\begin{figure}[t]
    \centering
    \includegraphics[width=0.7\linewidth]{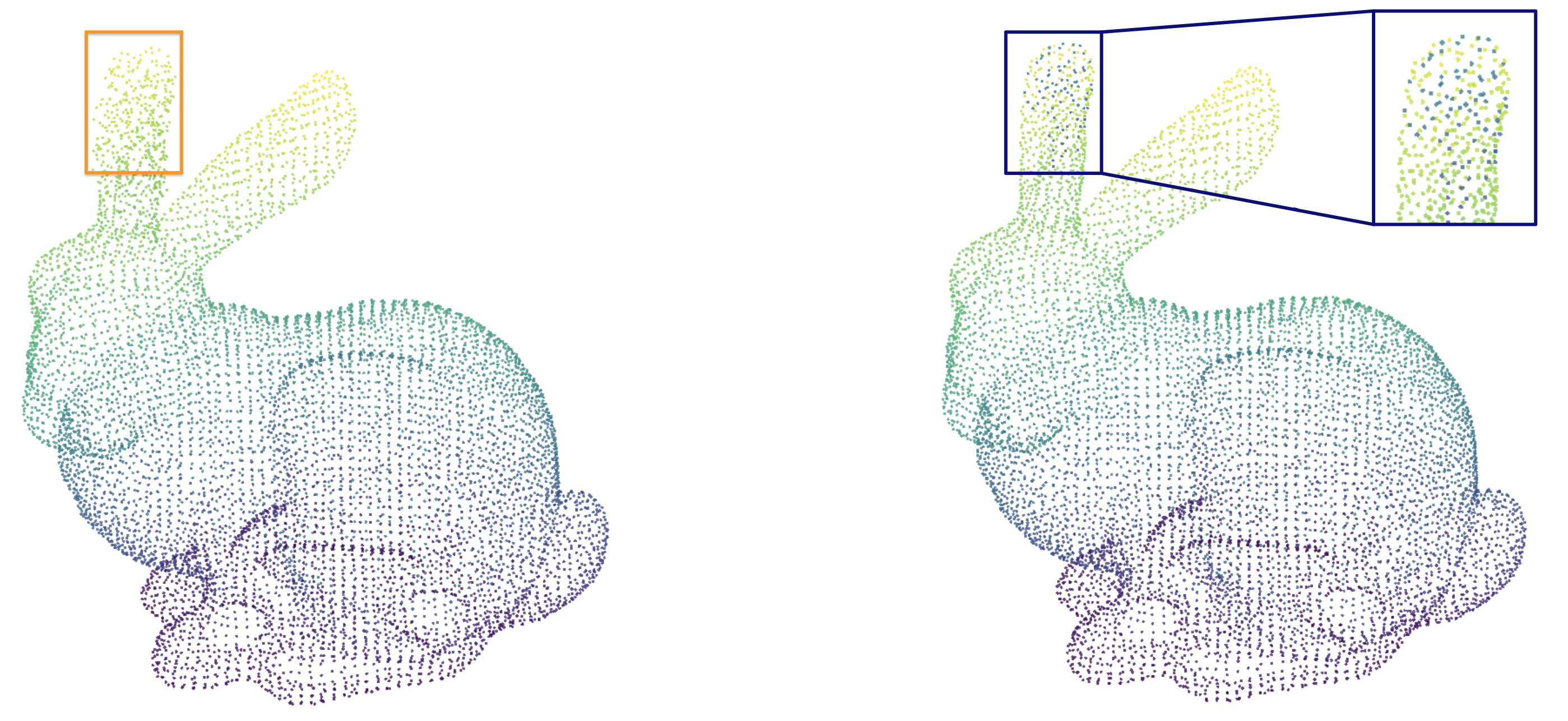}
    \caption{Shape and color defects: localized surface roughness ($\text{SNR} = 10.0$, left), incorrect material deposition (shift magnitude $5.0$, right).}
    \label{fig:bunny_oc}
\end{figure}

\subsection{In-Control and Out-of-Control Performance}\label{sec:sim-icoc}

\begin{table}[t]
\centering
\caption{Average ARL with standard deviation in parentheses for varying sample sizes $m_3$.
}
\label{tab:ic}
\resizebox{0.85\textwidth}{0.12\textheight}{
\begin{tabular}{lccccc}
    \toprule
    & \multicolumn{5}{c}{$m_3$} \\
    \cmidrule(lr){2-6}
    Method & 500 & 1,000 & 2,000 & 3,000 & 4,000\\
    \midrule
    {GL} & 122.49 (19.65) & 102.47 (4.94) & 96.92 (4.19) & 96.04 (4.08) & 95.27 (3.73)\\
    LLE &  178.47 (31.34) &  164.75 (29.40)  & 189.74 (43.90) & 181.88 (39.56) & 187.73 (44.38)\\
    ISOMAP & 131.47 (14.86) & 119.74 (9.44) & 102.85 (5.82) & 103.39 (5.36) & 100.85 (4.31) \\
    SMAC & 120.68 (18.96) & 105.57 (9.24) & 106.03 (6.29) & 101.42 (4.96) & 100.11 (4.32)
 \\
    \bottomrule
\end{tabular}
}
\end{table}

Table \ref{tab:ic} reports the attained IC AARL across varying sample sizes $m_3$. The $\text{ARL}_0$ is attained for SMAC and GL when $m_3 \geq$ 1,000. %However, although the AARL for GL stabilizes, it remains systematically below the target value. 
ISOMAP requires a larger sample size, $m_3 \geq$ 2,000, to attain $\text{ARL}_0$, whereas LLE fails to attain the target even at $m_3 = $ 4,000 and is therefore excluded from the OC comparison. In the following, we compare the detection performance of SMAC, GL, and ISOMAP using $m_3 = $ 4,000.
\begin{figure}[t]
    \centering
    \begin{subfigure}[b]{0.49\textwidth}
        \centering
        \includegraphics[width=\textwidth]{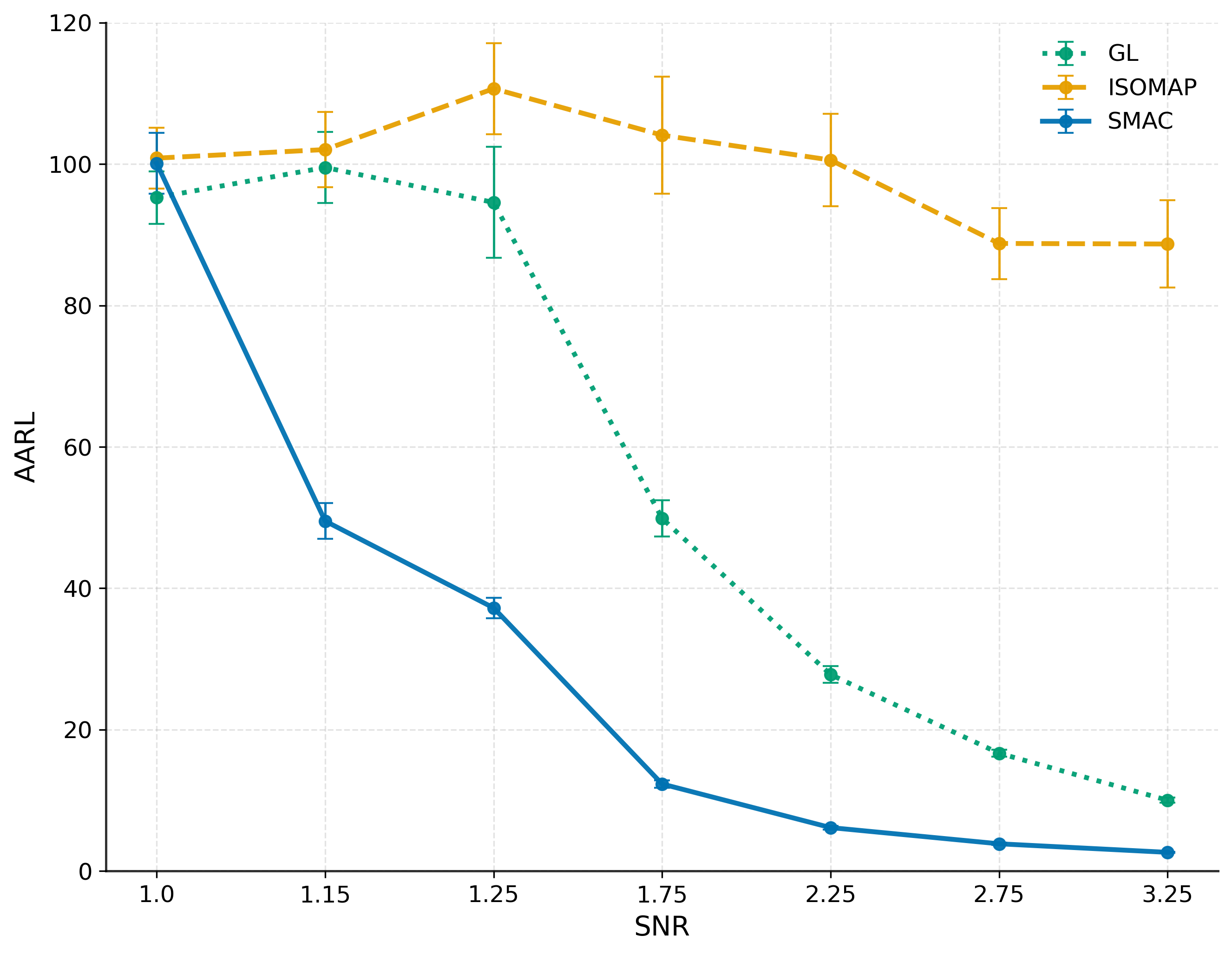}
        \caption{}
        \label{fig:bunny_shape}
    \end{subfigure}
    \hfill
    \begin{subfigure}[b]{0.49\textwidth}
        \centering
        \includegraphics[width=\textwidth]{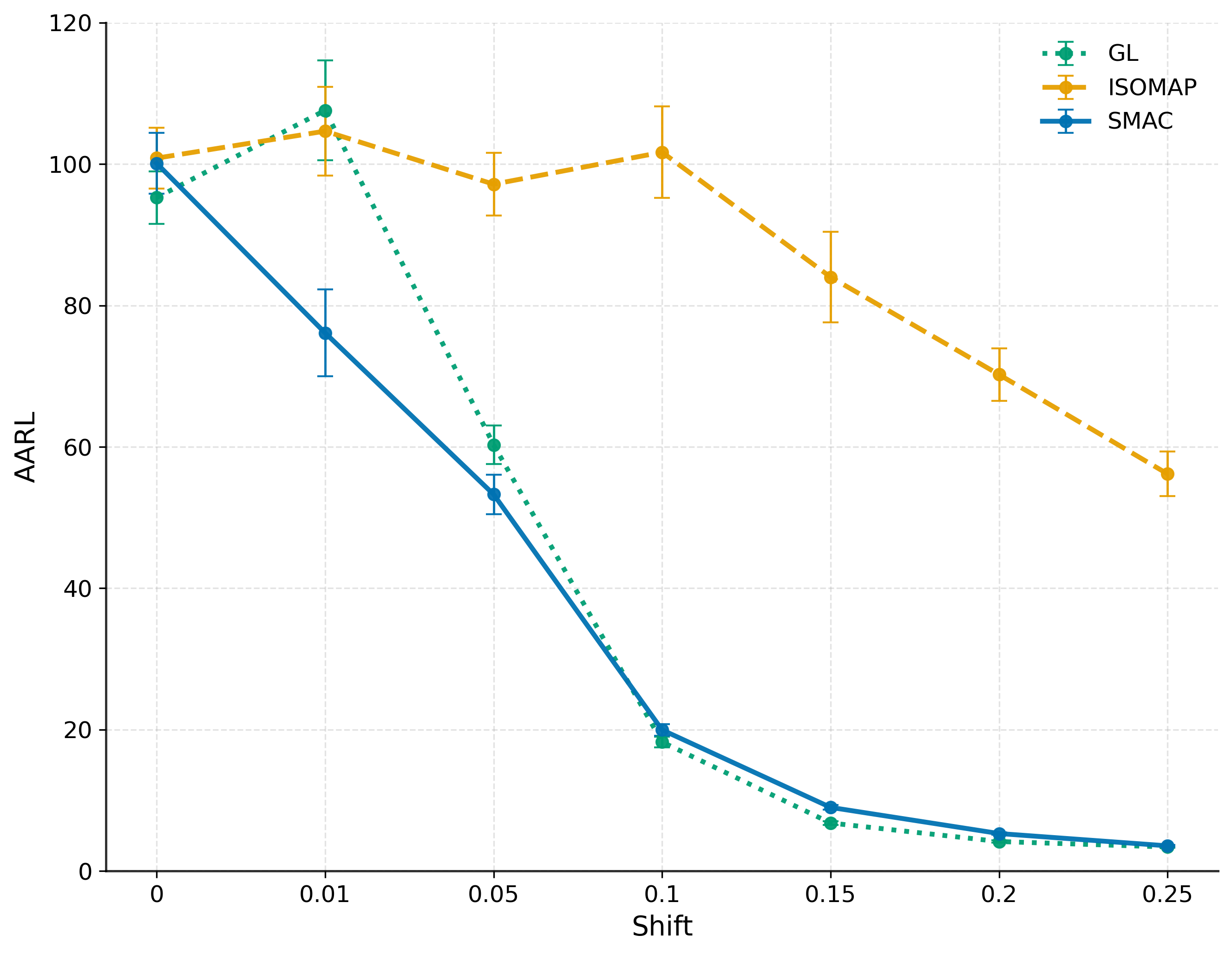}
        \caption{}
        \label{fig:bunny_color}
    \end{subfigure}
    \caption{Average ARL with standard deviation for competing methods under (a) shape change and (b) color change scenarios.}
    \label{fig:bunny_comparison}
\end{figure}

To assess the OC performance, two scenarios are first considered: a shape change and a color change (Figure \ref{fig:bunny_oc}). The results are reported in Figures \ref{fig:bunny_shape} and \ref{fig:bunny_color}.
\begin{figure}[htb!]
    \centering
    \begin{subfigure}[b]{0.49\textwidth}
        \centering
        \includegraphics[width=\textwidth]{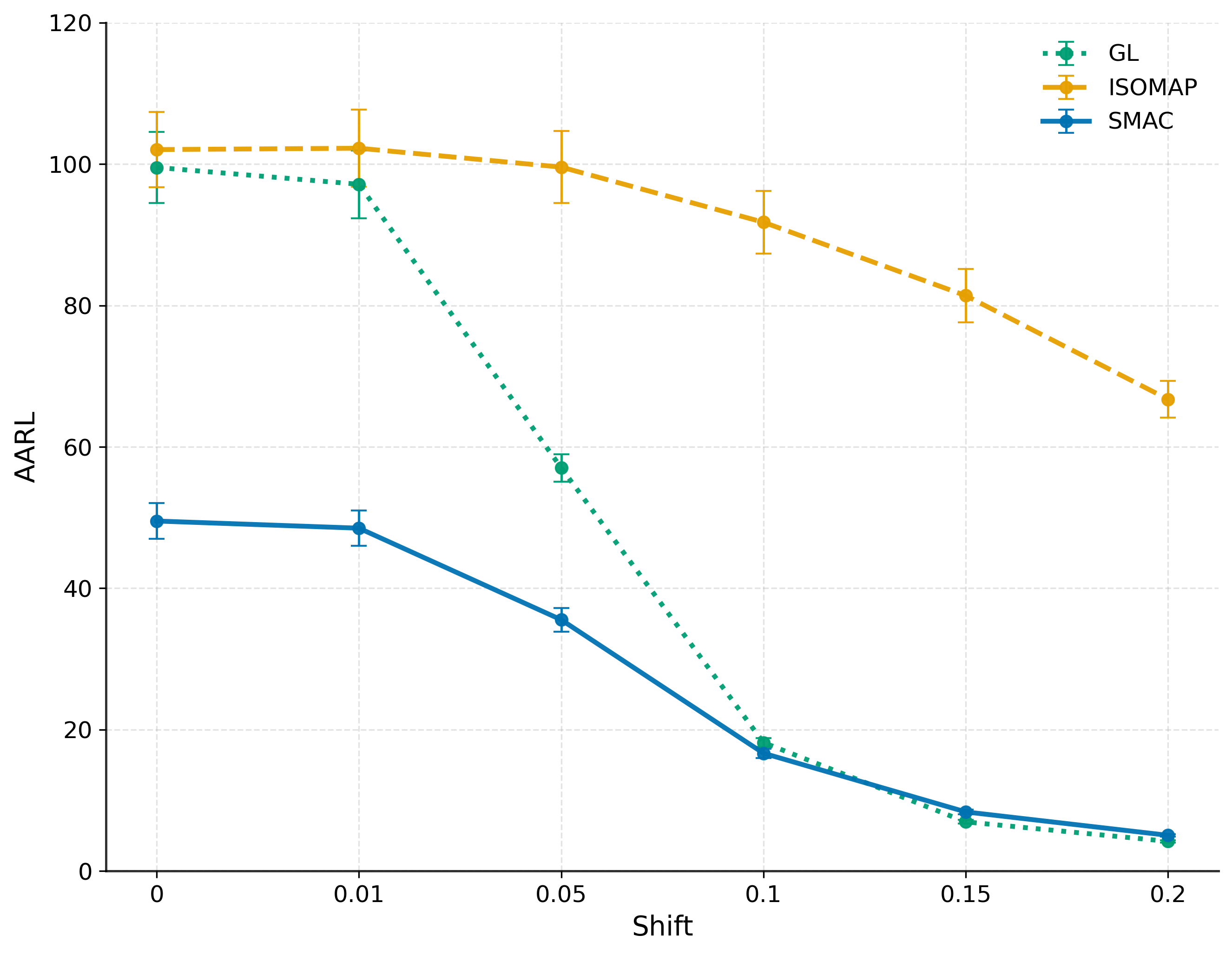}
        \caption{$\text{SNR} = 1.15$}
        \label{fig:bunny_cspot1}
    \end{subfigure}
    \hfill
    \begin{subfigure}[b]{0.49\textwidth}
        \centering
        \includegraphics[width=\textwidth]{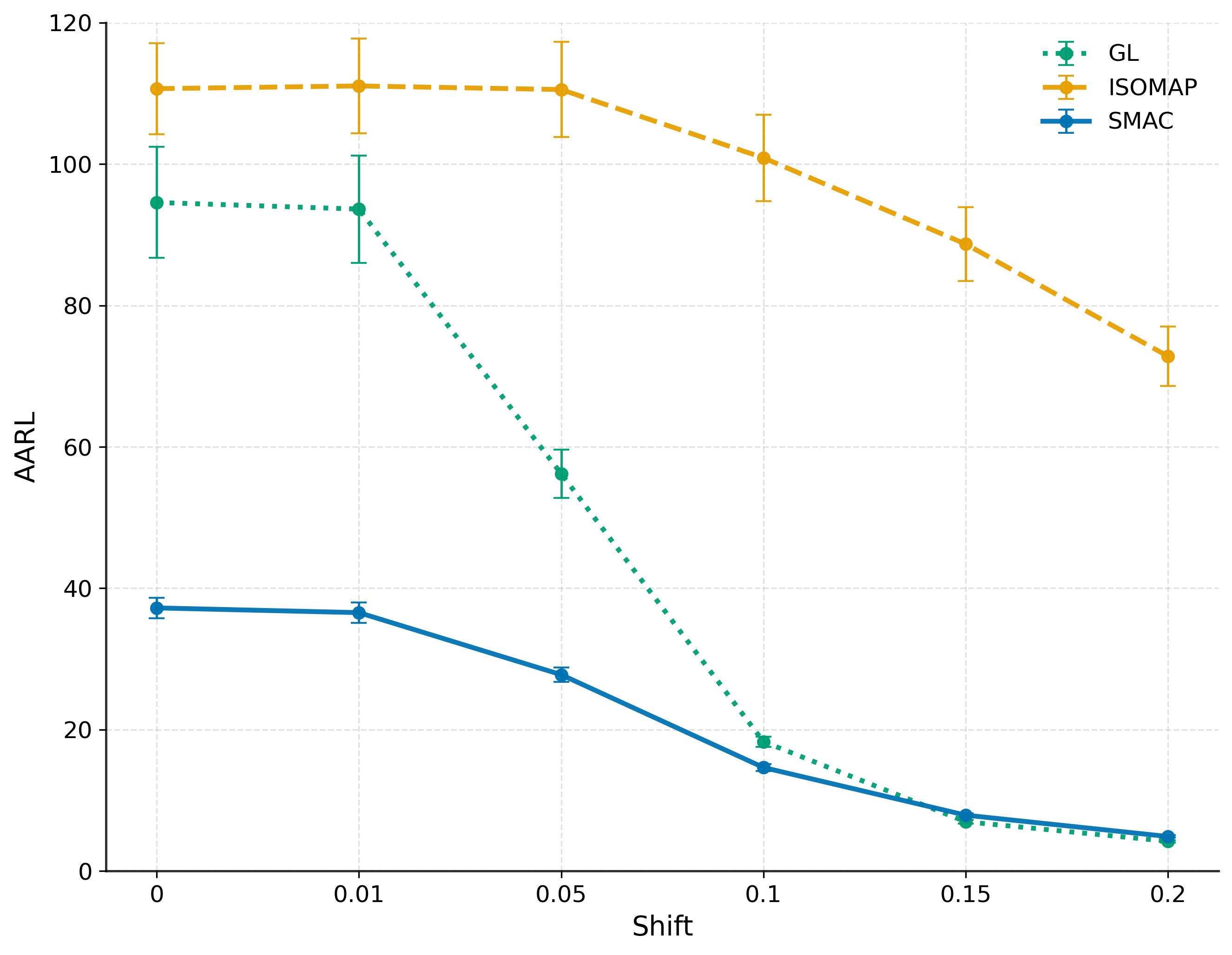}
        \caption{$\text{SNR} = 1.25$}
        \label{fig:bunny_cspot2}
    \end{subfigure}
    
    \vspace{1em} % Add vertical spacing between rows
    
    \begin{subfigure}[b]{0.49\textwidth}
        \centering
        \includegraphics[width=\textwidth]{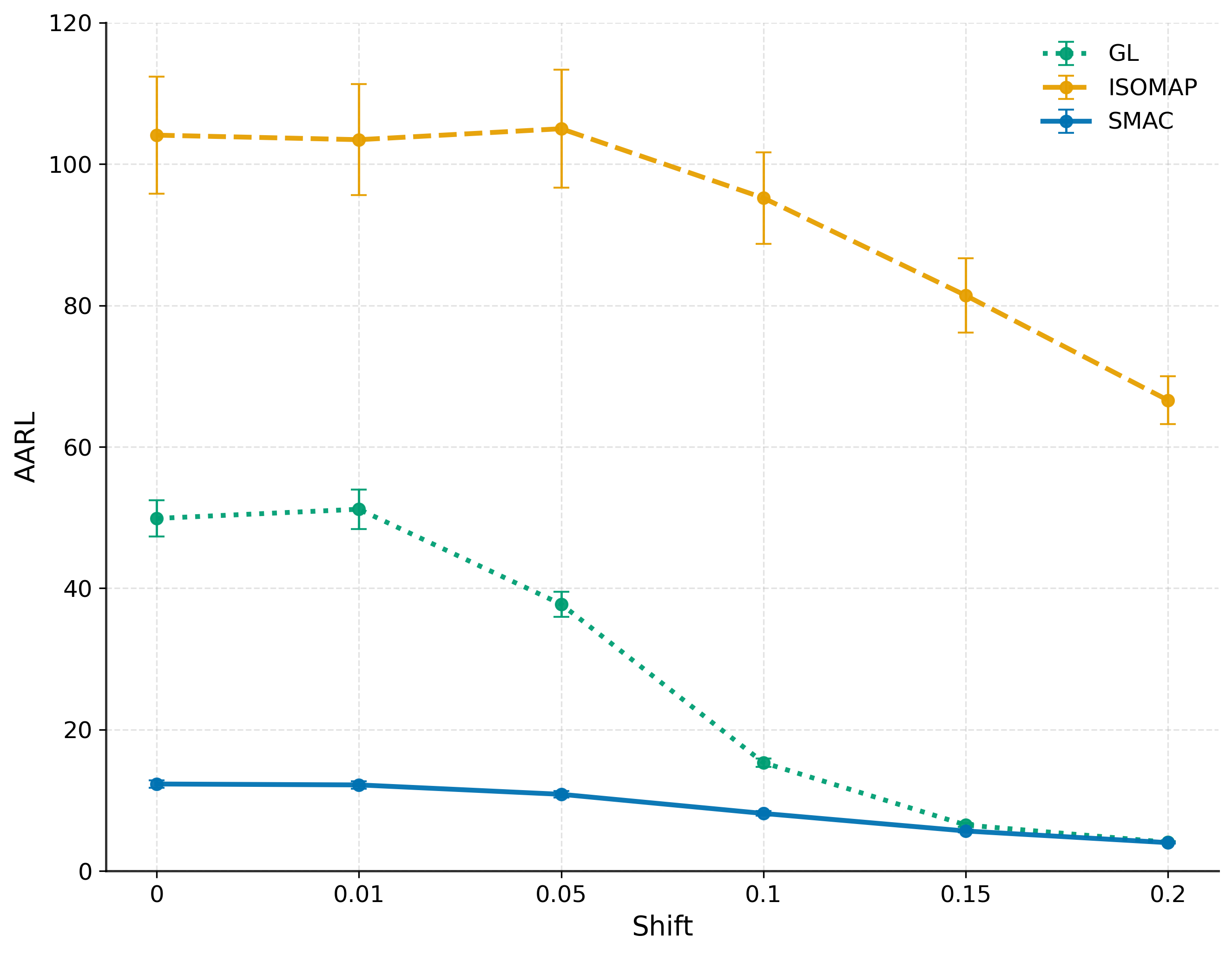}
        \caption{$\text{SNR} = 1.75$}
        \label{fig:bunny_cspot3}
    \end{subfigure}
    \hfill
    \begin{subfigure}[b]{0.49\textwidth}
        \centering
        \includegraphics[width=\textwidth]{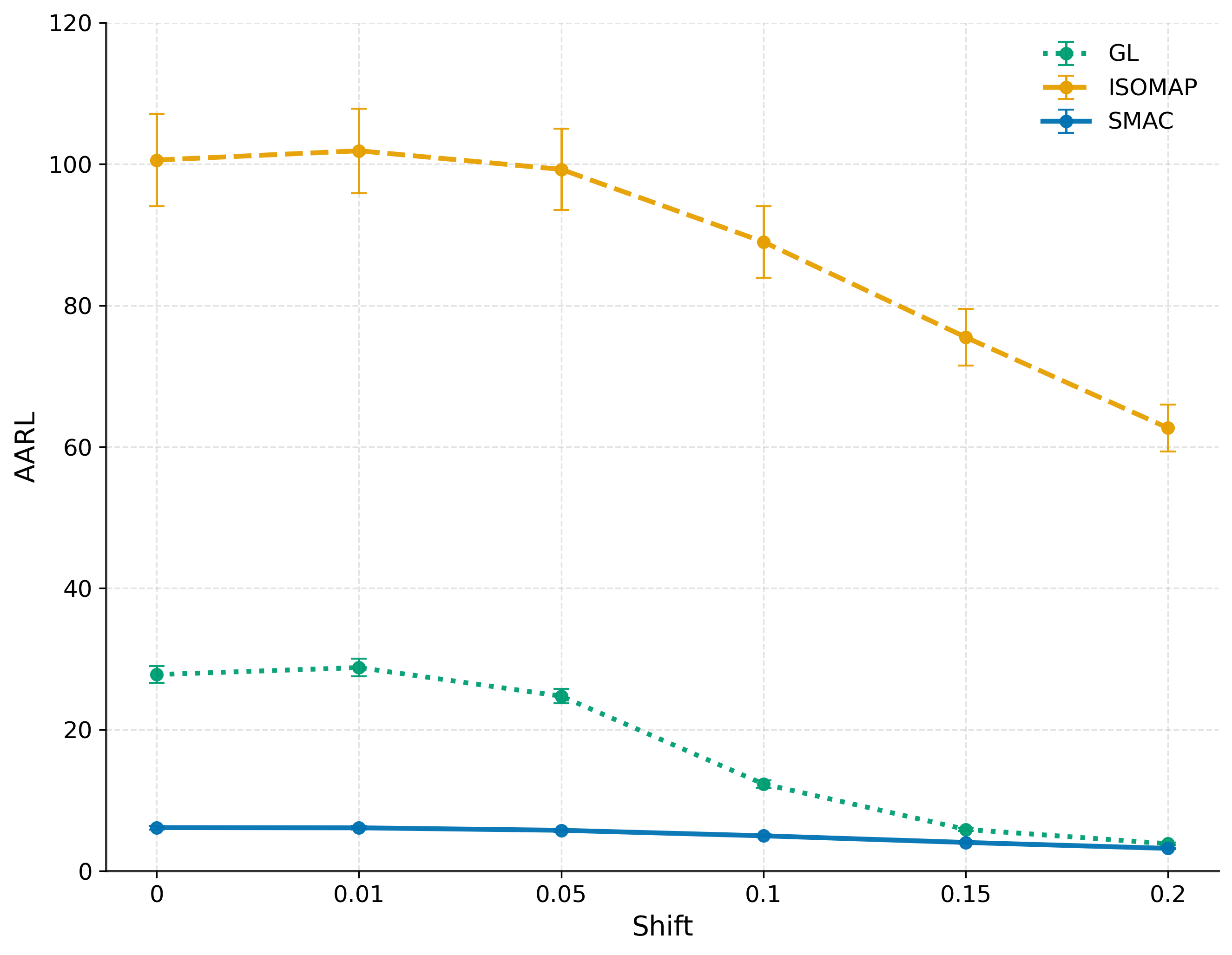}
        \caption{$\text{SNR} = 2.25$}
        \label{fig:bunny_cspot4}
    \end{subfigure}
    
    \caption{Average ARL with standard deviation for competing methods under simultaneous shape and color changes.}
    \label{fig:bunny_cspot}
\end{figure}
In the shape change scenario, localized surface roughness is simulated, applying isotropic noise sampled from $N_3(\bm{0}, \tilde{\tau}_s^2 \, \bm{I})$, with $\tilde{\tau}_s >  {\tau}_s$, to a small region. Results are reported in terms of signal-to-noise ($\text{SNR}$) ratio, defined as $\text{SNR}=\tilde{\tau}_s / {\tau}_s$. The affected region is a spatially clustered subset comprising approximately $5\%$ of the points located on one ear of the bunny (see Figure \ref{fig:bunny_oc}, left). Figure \ref{fig:bunny_shape} demonstrates the effectiveness of SMAC in detecting such a localized defect across the entire range of SNR values considered. The color change scenario represents incorrect material deposition, where the wrong material is deposited at specific locations. This defect is simulated by applying color shifts to isolated spots within the affected region (see Figure \ref{fig:bunny_oc}, right). These spots comprise approximately $1\%$ of the total points (distributed within the $5\%$ affected region mentioned earlier). %, with shift magnitudes ranging from 0.01 to 0.25.
Results presented in Figure \ref{fig:bunny_color} show that SMAC outperforms competing methods for smaller shift magnitudes and matches the best competing method (i.e., GL) for larger magnitudes. Remarkably, this superior performance is achieved despite SMAC operating at a disadvantage: SMAC uses a combined monitoring scheme where only the color chart can detect color changes, while competing methods employ dedicated single charts. This design means SMAC's color chart has an IC ARL nearly twice that of the competing methods, yet SMAC still demonstrates better or equivalent detection capability across all shift magnitudes. Despite the inherent disadvantage, the combined monitoring scheme provides a crucial diagnostic advantage: when a signal occurs, SMAC can identify whether the anomaly stems from shape changes, color changes, or both, whereas competing methods with single charts cannot distinguish the source of the defect.

Figure \ref{fig:bunny_cspot} presents the OC performance under simultaneous shape and color changes across four SNR levels ($\text{SNR} = 1.15, 1.25, 1.75, 2.25$). In the scenarios considered in Figure \ref{fig:bunny_cspot}, when the color shift is zero, the reported values correspond to a shape-only defect, with the SNR indicating the severity of the shape defect.  Across all scenarios, SMAC consistently outperforms competing methods for smaller color shifts and achieves a performance equivalent to GL for larger shifts, while ISOMAP demonstrates poor performance across all shift magnitudes. %This consistent pattern demonstrates SMAC's effectiveness in detecting combined defects. 

\subsection{Robustness to Variation in Point Density}\label{sec:sim-rob}

\begin{figure}[t]
    \centering
    \begin{subfigure}[b]{0.49\textwidth}
        \centering
        \includegraphics[width=\textwidth]{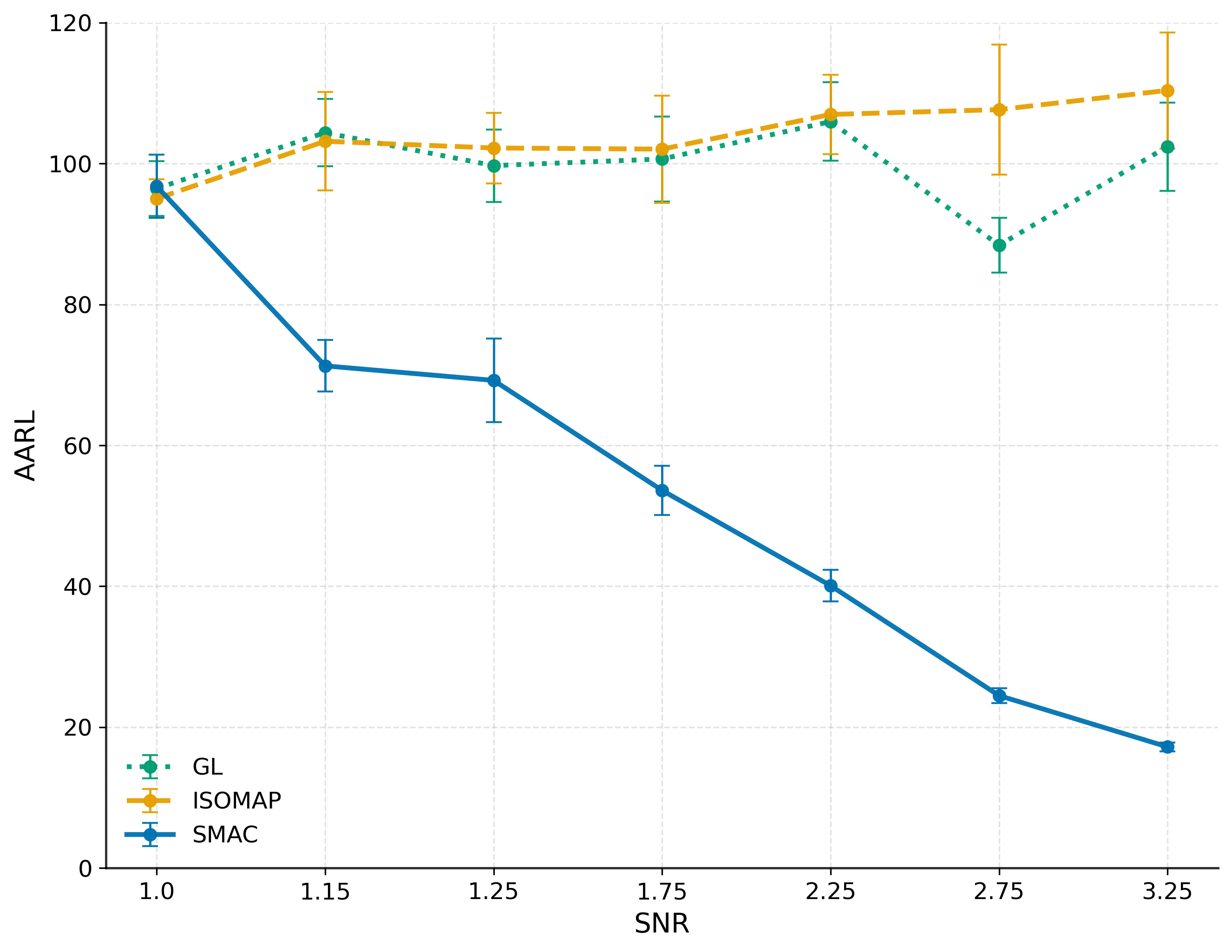}
        \caption{}
        \label{fig:bunny_shape_var}
    \end{subfigure}
    \hfill
    \begin{subfigure}[b]{0.49\textwidth}
        \centering
        \includegraphics[width=\textwidth]{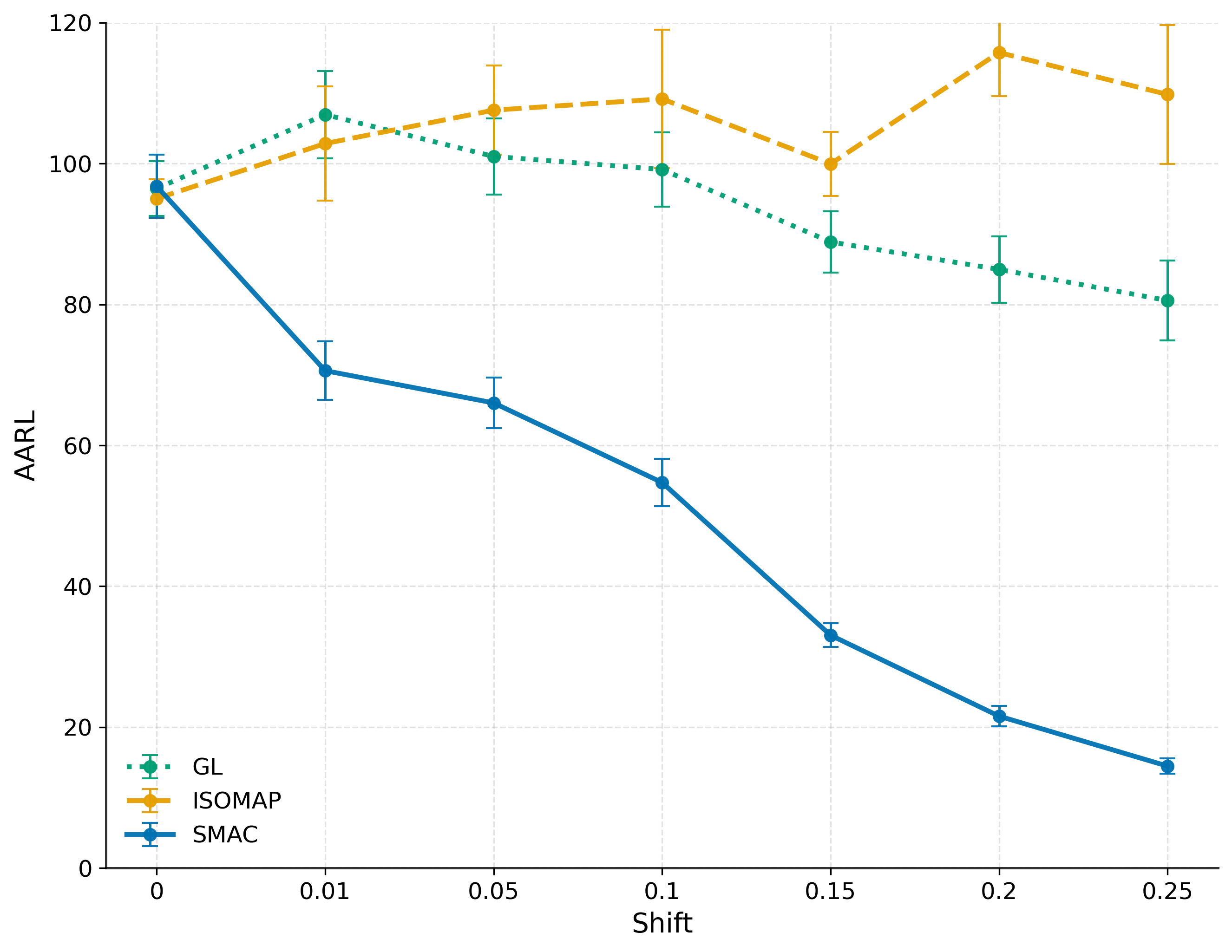}
        \caption{}
        \label{fig:bunny_color_var}
    \end{subfigure}
    \caption{Average ARL with standard deviation for competing methods under increased variation in point density: (a) shape change and (b) color change scenarios.}
    \label{fig:bunny_comparison_var}
\end{figure}

In this section, we assess the robustness of SMAC and competing methods to variation in point density. Specifically, noisy point clouds considered in this section contain between 6,480 and 8,140 points, while all other settings and tuning parameters are kept fixed.

Results for the shape change and the color change scenarios are reported in Figure \ref{fig:bunny_comparison_var}.
In both scenarios, SMAC exhibits superior performance to competing methods across all shift magnitudes, demonstrating its robustness in the presence of increased variation in point density. In contrast, GL and ISOMAP fail to detect any OC conditions, with AARL values close to the nominal ARL$_0$. 

\subsection{Post-signal Diagnostic Performance}\label{sec:sim-ps}
To evaluate the performance of the post-signal diagnostic procedure, we consider the shape alone change and combined shape and color changes. Results reported in the following refer to $\text{SNR}=1.25$ and color shifts ranging between 0.01 and 0.20 (Figure \ref{fig:bunny_cspot2}). The significance level $\alpha$ is set to 0.01 to control the FWER.
The number of point-wise descriptors $p$ is set to $200$ and $\eta = 0.001$, following the recommendations in \cite{HKS}, \cite{WKS} and \cite{Ovsjanikov2012}. Based on our empirical findings, the number of nearest neighbors for constructing the adjacency matrix used in the spatially-aware point-wise test is set to $10$.
\begin{figure}[t]
    \centering
    \begin{subfigure}[b]{0.49\textwidth}
        \centering
        \includegraphics[width=0.5\textwidth]{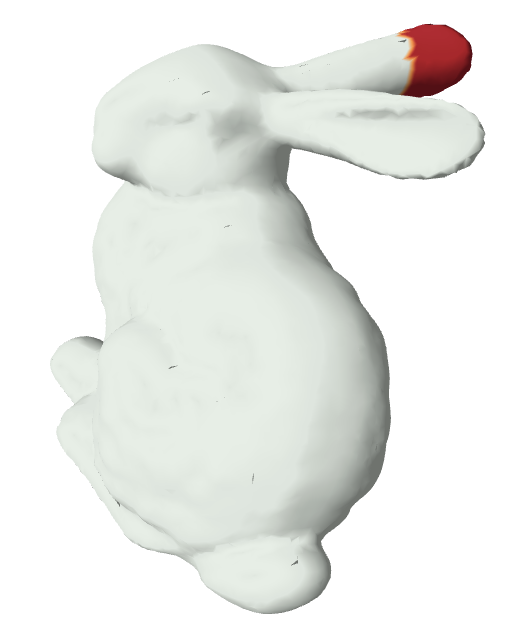}
        \caption{}
        \label{fig:bunny_mask}
    \end{subfigure}
    \hfill
    \begin{subfigure}[b]{0.49\textwidth}
        \centering
        \includegraphics[width=0.5\textwidth]{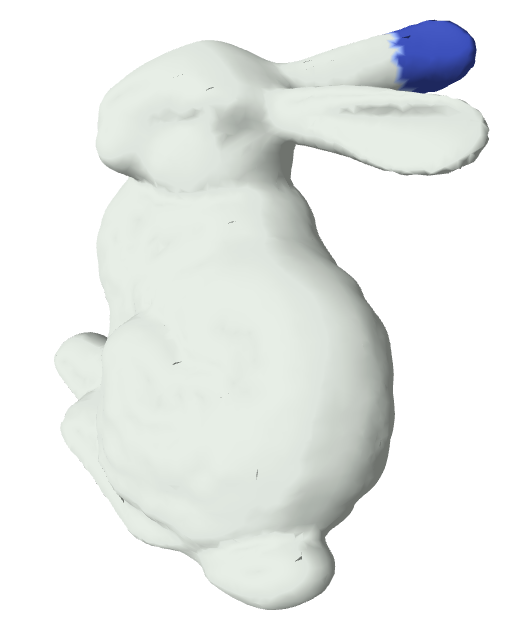}
        \caption{}
        \label{fig:bunny_mask_detected}
    \end{subfigure}
    \caption{Example of SMAC localization: (a) true anomalous region (color change) and (b) detected region using the spatially-aware point-wise test.
}
    \label{fig:bunny_ps}
\end{figure}

Table \ref{tab:diagnostic_equal} reports the diagnostic performance based on 100 simulation runs. Performance is evaluated using standard metrics: detection rate (DR), false positives (FP), precision, recall, F1-score, and intersection over union (IoU). The metrics are computed only on runs where anomalies were detected and, except for DR, results refer to averaged values across runs. In particular, results are presented for balanced group sizes of 300 samples. When only the shape change occurs, SMAC diagnostic correctly maintains the FWER at the nominal level $\alpha$, while the metrics related to anomaly localization are not reported since no color anomalous region is present.
As the color shift magnitude increases, the diagnostic performance improves substantially. At color shift 0.10, the detection rate reaches 84\% with an F1-score of 0.702. For color shifts of 0.15 and above, the detection rate reaches 100\%, while maintaining F1-scores above 0.83 and IoU values above 0.71. Figure \ref{fig:bunny_ps} shows an example of localization of the anomalous region. 
\begin{table}[t]
\centering
\caption{Diagnostic performance for shape alone and combined shape and color changes with balanced group sizes ($300$ samples in each group). Color defect applied to 1\% of points.}

\label{tab:diagnostic_equal}
\resizebox{0.68\textwidth}{0.15\textheight}{
\begin{tabular}{ccccccc}
\toprule
{Color Shift} & DR & FP & Precision & Recall & F1 & IoU \\
\hline
0.00 & 0.01 & 121.00 & / & / & / & / \\
0.01 & 0.07 & 187.86 & 0.00 & 0.00 & 0.00 & 0.00 \\
0.05 & 0.23 & 50.17 & 0.765 & 0.236 & 0.332 & 0.227  \\
0.10 & 0.84 & 25.92 & 0.899 & 0.613 & 0.702 & 0.567  \\
0.15 & 1.0 & 44.09 & 0.843 & 0.841 & 0.830 & 0.718  \\
0.20 &  1.0 & 65.74 & 0.766 & 0.932 & 0.835 & 0.722 \\
\hline
\end{tabular}
}
\end{table}
When considering smaller and unbalanced group sizes, the diagnostic performance deteriorates, especially for small color-shift magnitudes (see Supplementary Material Section S3). 
A simpler and effective approach for a single sample consists of thresholding the texture difference map: we compute the difference between the mean of the IC reconstructed textures and the OC reconstructed texture, then determine whether positive or negative differences exhibit larger absolute deviations and retain only the most extreme values in that direction. This simple approach provides a coarse localization of the anomalous region, but it lacks statistical guarantees (i.e., FWER control); see Section \ref{sec:casestudy} for an example of application.

\section{Case Study}\label{sec:casestudy}
This section presents the application of SMAC to a case study involving the design  of a multi-material bicycle seat described in Section \ref{sec:intro} (Figure \ref{fig:bike_sim}). The bicycle seat design combines two materials: Vero (rigid) and Agilus30 (soft) to mitigate deformations in unsupported regions while ensuring rider comfort (see \citealp{openvcad} for further details).

In this section, we validate the detection and diagnostic capabilities of SMAC in a real-world manufacturing example. Notably, our approach efficiently handles the high dimensionality of the point cloud through a sparse representation (see Supplementary Material Section S1 for time complexity analysis). In contrast, competing methods such as ISOMAP and LLE are computationally infeasible for HD point clouds due to their prohibitive memory requirements. Furthermore, as demonstrated in the simulation study, LLE fails to attain the target $\text{ARL}_0$, rendering it inadvisable regardless of computational constraints. Although GL employs a sparse representation, it relies on a single CUSUM chart, which prevents the decoupling of shape and color achieved by our approach. For these reasons, we omit the competing methods from the case study presented in this section. 

\begin{figure}[t]
    \centering    \includegraphics[width=0.5\linewidth]{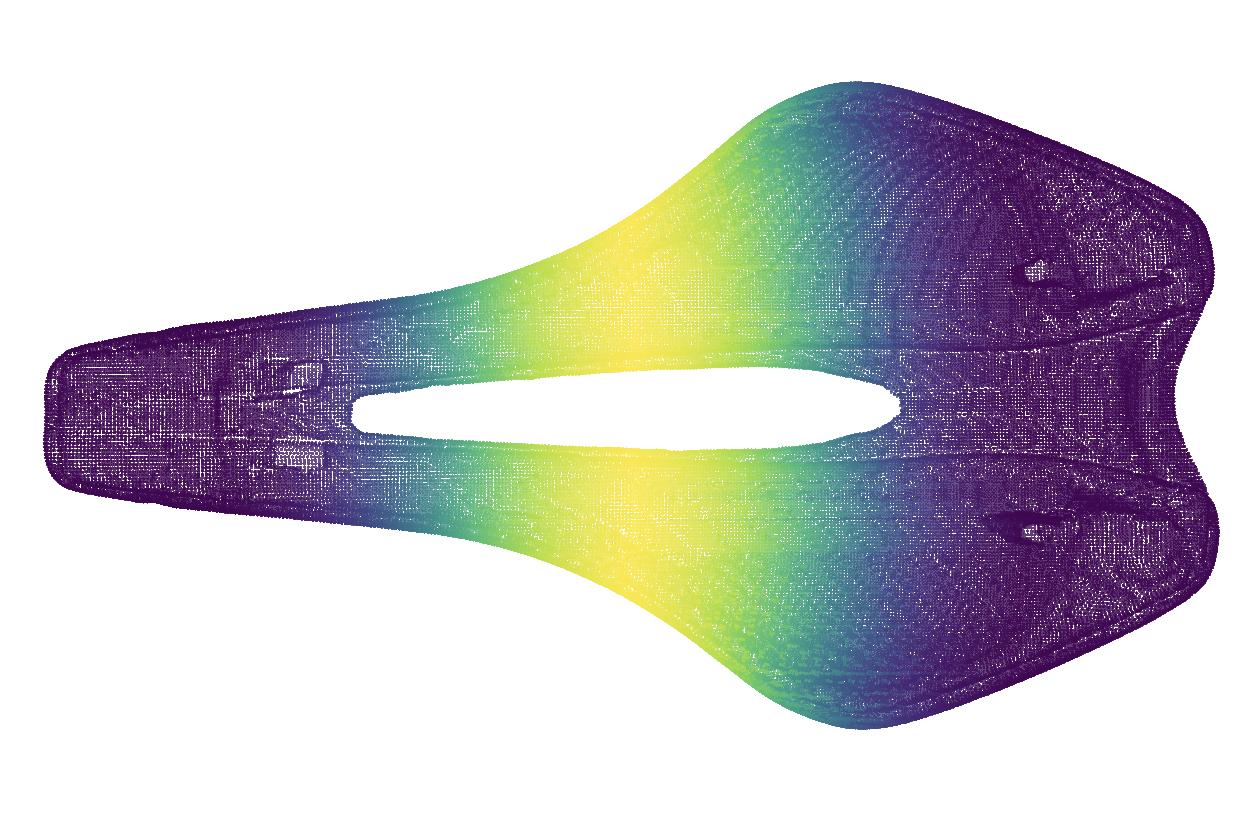}
    \caption{Point cloud of the bicycle seat: nominal design with chromatic attribute.}
    \label{fig:bike_case_study}
\end{figure}

The nominal design of the bicycle seat consists of a point cloud with approximately 93,000 points featuring spatially 
varying color, with color values ranging between 0 and 10 (Figure \ref{fig:bike_case_study}). Noisy point clouds of varying size are generated by randomly discarding points, adding isotropic noise $N_3(\bm{0}, \tau_s^2 \, \bm{I})$ with $\tau_s = 0.001$, and chromatic noise $N(0,\tau_c^2)$ with $\tau_c = 0.01$. Using the elbow rule to select $k$, the number of eigenvalues is set to $120$. The sample sizes $m_1$, $m_2$ and $m_3$ are set to 600, 300 and 1,000, respectively. The CUSUM reference values $k_s$ and $k_c$ are set to 0.05 and 1,000 bootstrap samples are used to compute the control limits. The control limit value for the shape chart is $h_s = 6.095$, while for the color chart $h_c = 4.771$. 

\begin{figure}[t]
    \centering
    \begin{subfigure}[b]{0.49\textwidth}
        \centering
        \includegraphics[width=\textwidth]{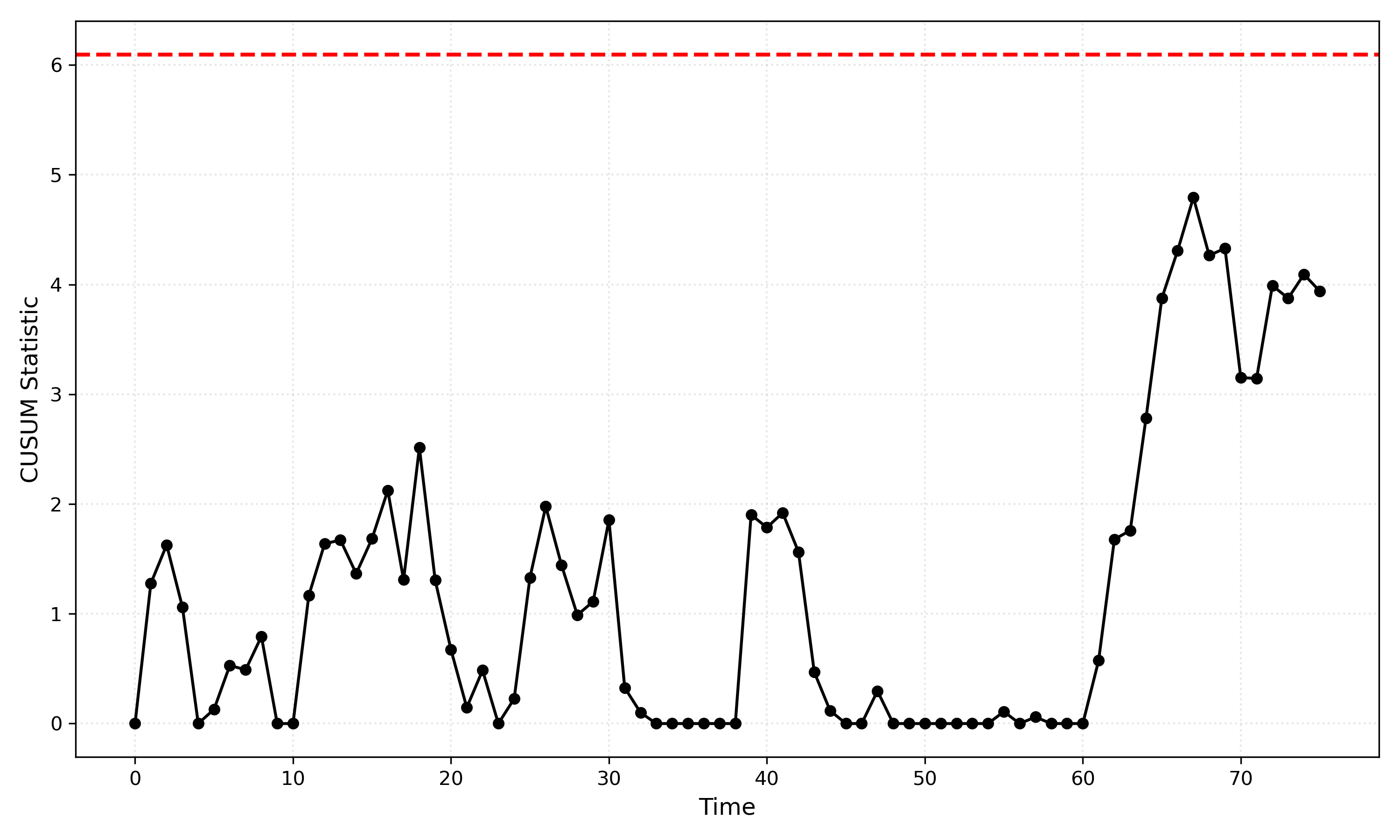}
        \caption{}
        \label{fig:shape_bike}
    \end{subfigure}
    \hfill
    \begin{subfigure}[b]{0.49\textwidth}
        \centering
        \includegraphics[width=\textwidth]{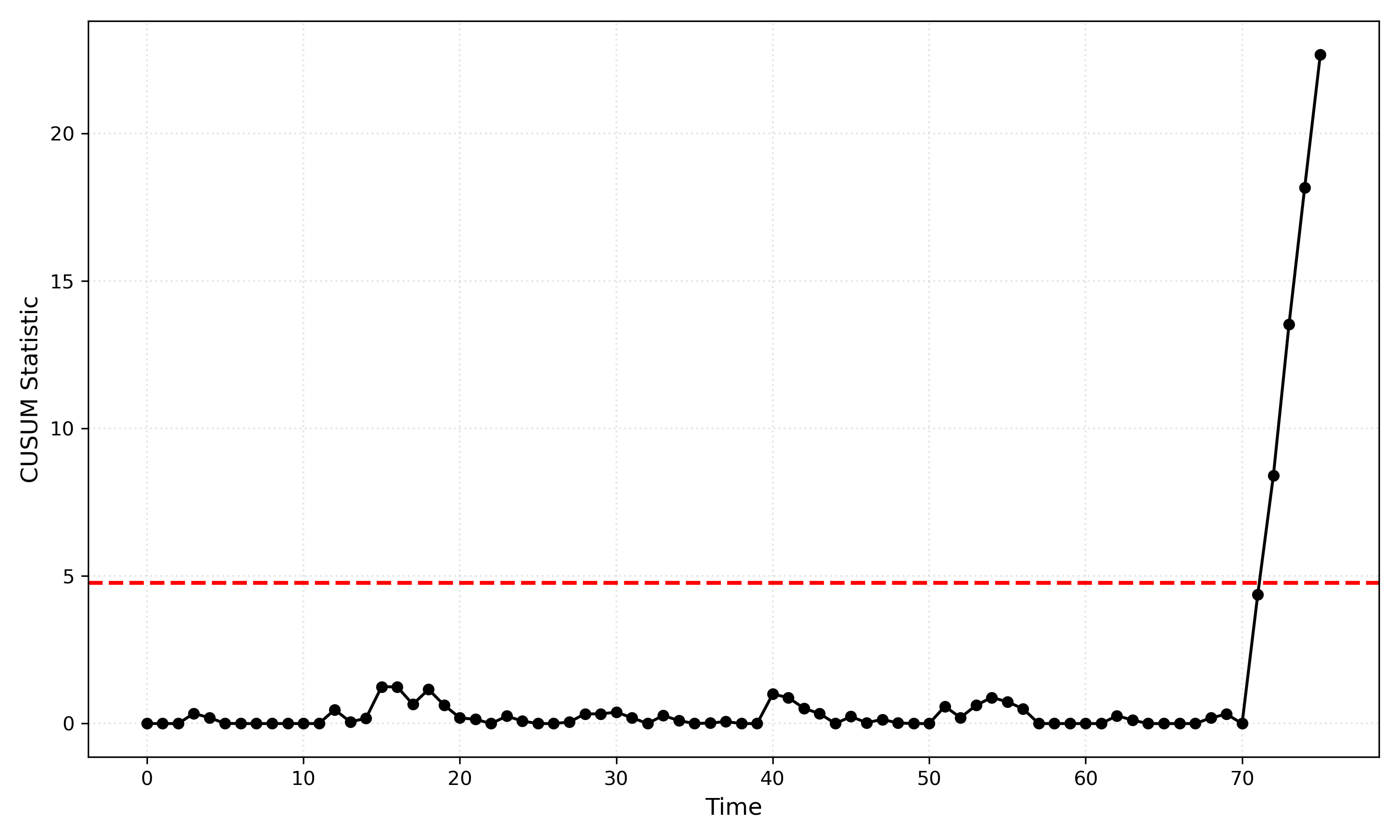}
        \caption{}
        \label{fig:color_bike}
    \end{subfigure}
    \caption{SMAC shape and color control charts: (a) and (b), respectively. The first seventy parts are IC, while the last five are OC.}
    \label{fig:case_study_bike}
\end{figure}

\begin{figure}[htb!]
    \centering
    \begin{subfigure}[b]{0.49\textwidth}
        \centering
        \includegraphics[width=0.48\textwidth]{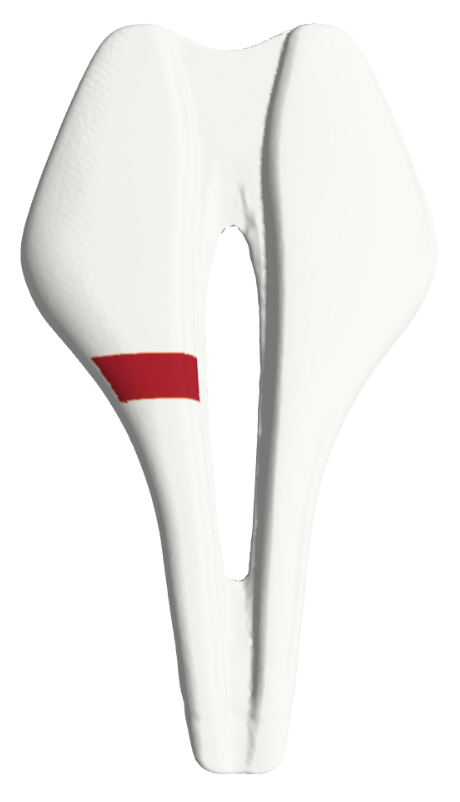}
        \caption{}
        \label{fig:bike_mask}
    \end{subfigure}
    \hfill
    \begin{subfigure}[b]{0.49\textwidth}
        \centering
        \includegraphics[width=0.48\textwidth]{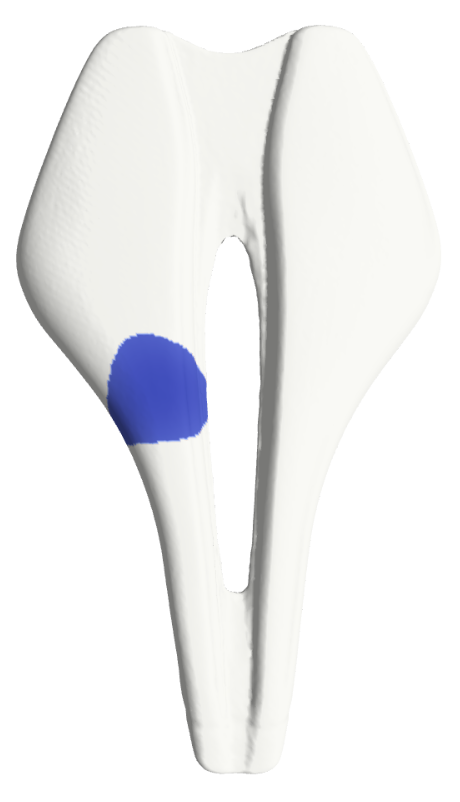}
        \caption{}
        \label{fig:bike_mask_detected}
    \end{subfigure}
    \caption{SMAC localization: (a) true anomalous region (color change) and (b) detected region using thresholding-based localization on a single OC sample.
}
    \label{fig:bike_ps}
\end{figure}

The OC scenario affects the color and represents incorrect material deposition. Specifically, within a localized region of the surface, 1\% of the points exhibit a darker color than the nominal (a downward shift of 0.2). The affected region is highlighted in red in Figure \ref{fig:bike_mask}. The sample includes 70 IC parts, followed by 5 OC parts. Figure \ref{fig:case_study_bike} depicts the shape and color control charts: SMAC's color chart triggers an alarm at $t = 72$, correctly identifying the change at the second OC part, while the shape chart does not trigger any alarms. 
However, in this scenario, the power of the spatially-aware point-wise test is limited since only a few OC parts are available for localization. Hence, we apply the thresholding-based localization approach instead. Figures \ref{fig:bike_mask} and \ref{fig:bike_mask_detected} show the true anomalous region and the localization obtained from a single OC part.

%% file: conclusion.tex
\section{Conclusions}\label{sec:conclusion}
Simultaneous monitoring of shape and surface color is particularly relevant in applications involving printed parts featuring spatially varying material composition, such as functionally graded materials. This paper presents a novel approach for monitoring raw, unaligned, and unstructured point clouds, augmented with point-wise color information via data fusion. The proposed approach, named SMAC, avoids preprocessing steps commonly adopted to simplify the analysis, such as alignment to a common orientation (registration) or mesh reconstruction to enable the application of established tools (e.g., finite element analysis). 

A Monte Carlo simulation study was conducted to assess the detection performance of SMAC and competing methods when subtle, localized defects occur. SMAC achieves effective detection of shape-only, color-only, and combined shape and color defects, while remaining robust to point density variation. Furthermore, a post-signal diagnostic procedure is presented since the monitoring scheme is inherently coupled. In particular, SMAC's diagnostic procedure aims to identify the source of the defect and localize color anomalies. The performance of the diagnostic procedure improves substantially as the magnitude of the color shift increases, providing accurate localization, as demonstrated by high F1-score and IoU values.
In addition, a case study on a functionally graded bicycle seat demonstrates its applicability in a real-world manufacturing setting. 

The main limitation of SMAC is that the diagnostic procedure may have low power when few OC parts are available. In such cases, we provided an alternative thresholding-based approach that achieves a rough localization of color anomalies, although it does not provide any FWER control. 
Another limitation is the relatively large number of real or synthetic IC samples required for calibration. This requirement, however, is closely related to the dimensionality of the extracted features $k$: a lower $k$ reduces the number of IC samples required for calibration, but at the cost of reduced detection accuracy. Therefore, selecting $k$ involves a trade-off between calibration sample size and detection performance. % which should be tuned according to the data availability and monitoring requirements of the specific application.
Additionally, SMAC assumes that 4D point clouds are obtained via data fusion and does not take into account potential alignment errors at this stage. 

Future developments include improving the power of the post-signal diagnostic procedure, so that it can operate effectively with few OC samples while controlling the FWER. Another important direction is the development of a monitoring framework that requires minimal IC samples for calibration.